\documentclass{sig-alternate}

\usepackage{epstopdf}
\usepackage{amsfonts}
\usepackage{subfigure}
\usepackage{amsmath}
\usepackage{comment}
\usepackage{algorithmic}
\usepackage{algorithm}
\usepackage{xcolor}
\usepackage{url}

\newcommand{\calX}{\mathcal{X}}
\newcommand{\bbR}{\mathbb{R}}

\usepackage{etoolbox}
\patchcmd{\maketitle}{\@copyrightspace}{}{}{}

\begin{document}





%

\clubpenalty=10000 
\widowpenalty = 10000

\title{Derivative Delay Embedding: Online Modeling of Streaming Time Series}

%
%
%
%
%

\numberofauthors{4} 
%
\author{ 
%
%
\alignauthor
Zhifei~Zhang\\
\affaddr{University of Tenneessee}\\
\affaddr{Knoxville, TN USA}\\
\email{zzhang61@vols.utk.edu}
\alignauthor
Yang~Song\\
\affaddr{University of Tenneessee}\\
\affaddr{Knoxville, TN USA}\\
\email{ysong18@vols.utk.edu}
\alignauthor
Wei~Wang\\
\affaddr{University of Tenneessee}\\
\affaddr{Knoxville, TN USA}\\
\email{wwang@vols.utk.edu}
\and  
\alignauthor
Hairong~Qi\\
\affaddr{University of Tenneessee}\\
\affaddr{Knoxville, TN USA}\\
\email{hqi@utk.edu}
}

\maketitle
\begin{abstract}
	The staggering amount of streaming  time series coming from the real world calls for more efficient and effective online modeling solution. For time series modeling, most existing works make some unrealistic assumptions such as the input data is of fixed length or well aligned, which requires extra effort on segmentation or normalization of the raw streaming data. Although some literature claim their approaches to be invariant to data length and misalignment, they are too time-consuming to model a streaming time series in an online manner.  
	We propose a novel and more practical online modeling and classification scheme, DDE-MGM, which does not make any assumptions on the time series while maintaining high efficiency and state-of-the-art performance. The derivative delay embedding (DDE) is developed to incrementally transform time series to the embedding space, where the intrinsic characteristics of data is preserved as recursive patterns regardless of the stream length and misalignment. Then, a non-parametric Markov geographic model (MGM) is proposed to both model and classify the pattern in an online manner. Experimental results demonstrate the effectiveness and superior classification accuracy of the proposed DDE-MGM in an online setting as compared to the state-of-the-art.
\end{abstract}

\printccsdesc



\vspace{5mm}
\noindent \textbf{\large Source code}
 
\url{https://github.com/ZZUTK/Delay_Embedding.git}

\newpage
\section{Introduction}
There has been an unprecedented surge of interest in streaming time series modeling and classification mainly due to the rapid deployment of smart devices. Traditionally, time series classification has been conducted using an offline training procedure coupled with an online/offline classification procedure. During the training process, some preprocessing steps are usually conducted including segmenting the time series into finite (usually fixed) length and aligning the segments perfectly to facilitate the subsequent feature extraction that normally yields discriminative patterns for classification purpose. However, with the smart device explosion and the related jump in data traffic, new challenges arise in time series data analysis. For example, time series data generally exhibit time-varying characteristics over an, in theory, infinite time span, therefore, manually truncating the time series into fixed-length, well-aligned segments would run the risk of missing some intrinsic characteristics of the data that degrades the performance of the classifier. On the other hand, many smart devices require real-time responses. Therefore, the computational complexity becomes the bottleneck for most classifiers. 

These challenges call for a solution that is able to model a time series in an online manner without the need for any preprocessing such that time-varying characteristics of the time series can be well captured in real time and that the classification can be performed using the most updated model. Existing works either are time-consuming or have to make certain assumptions on the times series, e.g.,  fixed length (as opposed to random or infinite length) and well alignment (i.e., the time series are aligned to the same starting point or baseline), which have largely hindered the realization of online modeling or classification. 

To the best of our knowledge, there has not been any related work that can achieve online processing in both modeling and classification stages without the assumptions of fixed length and well alignment. We develop the derivative delay embedding (DDE) method that transforms, in an online fashion, a time series to the embedding space, in which the patterns are preserved regardless of the assumptions mentioned above. We further propose the Markov geographic model (MGM) that enables both the modeling and classification of the transformed patterns in an online fashion. We refer to the proposed approach as DDE-MGM.

\subsection{Motivation}
The theory of delay embedding~\cite{takens1981detecting,sauer1991embedology} was first introduced to reconstruct a chaotic dynamical system from a sequence of observations of the system. The reconstruction preserves the coordinate and period changes of the dynamical system, but it is invariant to the change of phase. Therefore, a time series can be considered a sequence of observations from a latent dynamical system, and we can represent the time series by the reconstructed dynamical system, which is invariant to phase changes (i.e., misalignment). Another merit of delay embedding is its low computational complexity, approximately $O(1)$. In addition, the reconstruction is performed in an incremental fashion, facilitating online processing, because only recent observations are considered when reconstructing the dynamical system from a streaming time series. The reconstructed dynamical system is usually represented in a higher dimensional space, in which the dynamics presents recursive patterns~\cite{kennel1992determining,perea2013sliding,lainscsek2015delay} regardless of the length of the original time series. Therefore, an infinite streaming time series can potentially be stored in a finite memory through the delay embedding because of the recursiveness of the reconstructed dynamical system. 

Motivated by the invariance properties of delay embedding, especially the invariance to the phase and length changes of the time series, we develop the online modeling and classification scheme, DDE-MGM, taking advantage of the invariance properties and high efficiency from the delay embedding technique.   

\subsection{Related Work}
The dynamic time warping (DTW) method~\cite{berndt1994using} has achieved good performance in time series classification, especially the 1NN-DTW~\cite{xi2006fast}. However, DTW-based methods normally suffer from the high computational complexity that is not suitable for many real-time applications. Several recent improvements~\cite{xi2006fast,petitjean2014dynamic} have successfully reduced the computational complexity, however, they are still far from achieving online processing. Other methods such as HMMs~\cite{lv2006recognition}, decision tree~\cite{rodriguez2004interval}, SVM~\cite{wu2004distance}, and neural network~\cite{hanson2009neural}, are also limited by their high computational complexity in the training stage and the necessity to make the two impractical assumptions, i.e., fixed length and well alignment of the time series. 

Recently, some works have been proposed attempting to relax these assumptions. For example, \cite{hu2013time} removed the assumption of fixed length by learning a dictionary, but it needs a long time to learn an appropriate dictionary. \cite{li2014robust,song2015multiple} removed the assumption of well alignment by sparse coding. However, they still have to learn a dictionary in an offline manner.
\cite{perea2013sliding,zhifei2015early} exploited the delay embedding technique in time series analysis, which relaxed both assumptions of fixed length and well alignment, but neither can be performed in the ``online'' scenario.  


Many online learning methods were also proposed in resent years, e.g.,  
\cite{orabona2008projectron} 
proposed the kernel based perceptron with budget, \cite{wang2010online} improved the online passive-aggressive algorithm, and 
\cite{lularge2015} 
extended online gradient descent. However, they require the data to be of the same length or well aligned. In addition, they are more suitable to operate in the feature space rather than on the raw time series. Therefore, we consider these methods as pseudo-online because they need to preprocess (i.e., truncating or aligning) the raw time series.






In this paper, we specifically consider the problem in a more practical ``online'' setting, where a time series streams in with random length. 

The contribution of this paper is three-fold. 
First, the proposed DDE completely removes the common but unrealistic assumptions made on the time series, i.e., fixed length and well alignment, therefore, no preprocessing is needed on the time series, facilitating real-world problem solving. 
Second, the proposed MGM effectively and efficiently models the trajectory of the recursive patterns of different classes in the embedding space. Thus, both the modeling and classification using DDE-MGM are performed in an online and incremental fashion, while maintaining competitive classification accuracy as compared to the state-of-the-art.
Third, the discretization of the embedding space enables an approximately constant memory footprint during the modeling regardless of the stream length.

\section{Derivative Delay Embedding}
\label{sec:Derivative Delay Embedding}
In this section, we first present background of delay embedding. Then, the proposed derivative delay embedding (DDE) is elaborated, as well as its invariant property to data length and misalignment. Finally, parameter selection for delay embedding and DDE is discussed to facilitate real-world applications of DDE.

\subsection{Delay Embedding}

A time series $[y_t,y_{t+1},\cdots]$ can be considered as an observable sequence from a latent deterministic dynamical system~\cite{laur2004timeseries}, which evolves in time
\begin{equation}
x_{t+1}=\phi(x_t),\;t=0,1,2,\cdots,
\label{eq:evolve}
\end{equation}
where $x_t\in\calX$ denotes the system state at time $t$, and $ \phi:\calX\rightarrow\calX $ is a deterministic function. Without loss of generality, we assume the time series is of high dimension, i.e., $y_t\in\bbR^n$. Because we cannot directly observe those internal states $x_t$ of the system, the states are measured via an observation function $\psi:\calX\rightarrow\bbR^n$. For each set of states $[x_t,x_{t+1},\cdots]$, there is a corresponding time series
\begin{align}
[y_t,y_{t+1},\cdots] & =\left[ \psi(x_t),\psi(x_{t+1}),\cdots \right]\\
~ & =\left[ \psi(x_t),\psi(\phi(x_t)),\cdots \right].
\end{align}

Our purpose is to estimate the deterministic function $\phi$ of the latent dynamical system by reconstructing the internal states $[x_t,x_{t+1},\cdots]$ from the observations $[y_t,y_{t+1},\cdots]$. For classification purpose, the times series of the same class should share similar $\phi$.

From Takens' embedding theory~\cite{takens1981detecting}, a series of observations need to be considered to reconstruct a single state because a state of the deterministic dynamical system is associated with current and recent observations. Assuming  $\calX$ is a smooth manifold, $\phi\in C^2$ is a diffeomorphism, and $\psi\in C^2$, the mapping $\Phi_\psi:\calX\rightarrow\bbR^n\times\bbR^d$, defined by
\begin{align}
\Phi(x_t;s,d)&=\left( \psi(x_t), \psi(x_{t+s}),\cdots,\psi(x_{t+(d-1)s}) \right)\\
&=\left( y_t, y_{t+s},\cdots,y_{t+(d-1)s} \right),
\label{eq:Phi}
\end{align}
is a \textit{delay embedding}, where $\Phi(x_t;s,d)$ is the reconstruction of the state $x_t$ in the Euclidean space, which is referred to as the \textit{embedding space}. The parameter $s$ is the delay step, and $d$ denotes the embedding dimension. Based on the reconstructed states $[\Phi(x_t),\Phi(x_{t+1}),\cdots]$, the deterministic function $\phi$ can be estimated. For simplicity, we use $\Phi(x_t)$ to denote $\Phi(x_t;s,d)$ in the rest of this paper. 

\begin{figure}[h]
	\centering
	\includegraphics[width=.8\columnwidth, trim=10 5 15 10, clip]{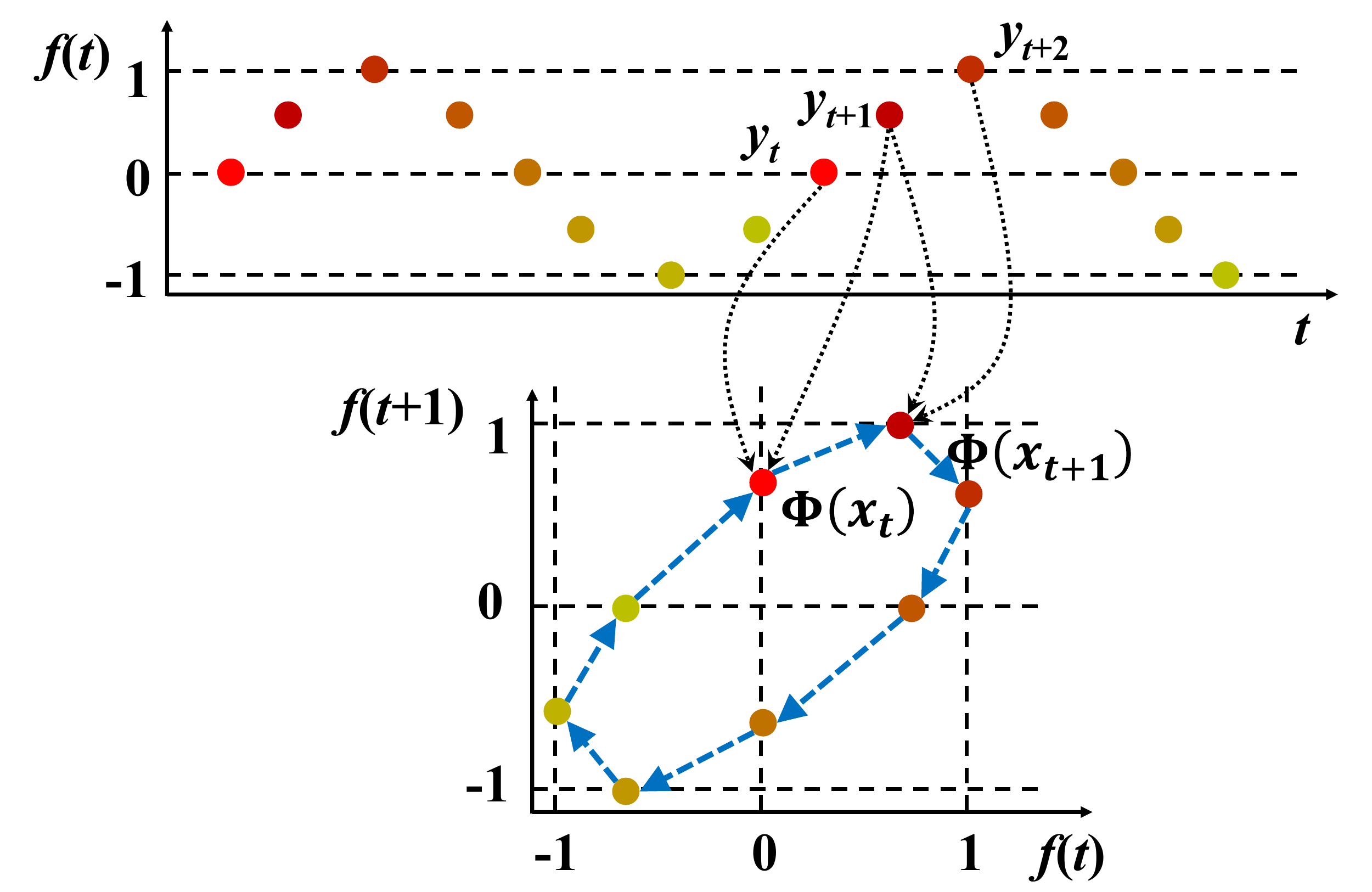}
	\caption{A toy example of delay embedding ($d=2$, $s=1$). Top: a time series. Bottom: states reconstructed from the time series through delay embedding. The black dotted arrows indicate the corresponding points in the time and embedding space. The dashed blue arrows show the trajectory that the states are constructed in the embedding space.}
	\label{fig:DE}
\end{figure}

\begin{figure}[h]
	\centering
	\subfigure[Time series]{\label{subfig:DE_comp_a}
		\includegraphics[width=.45\columnwidth, trim=5 5 20 15, clip]{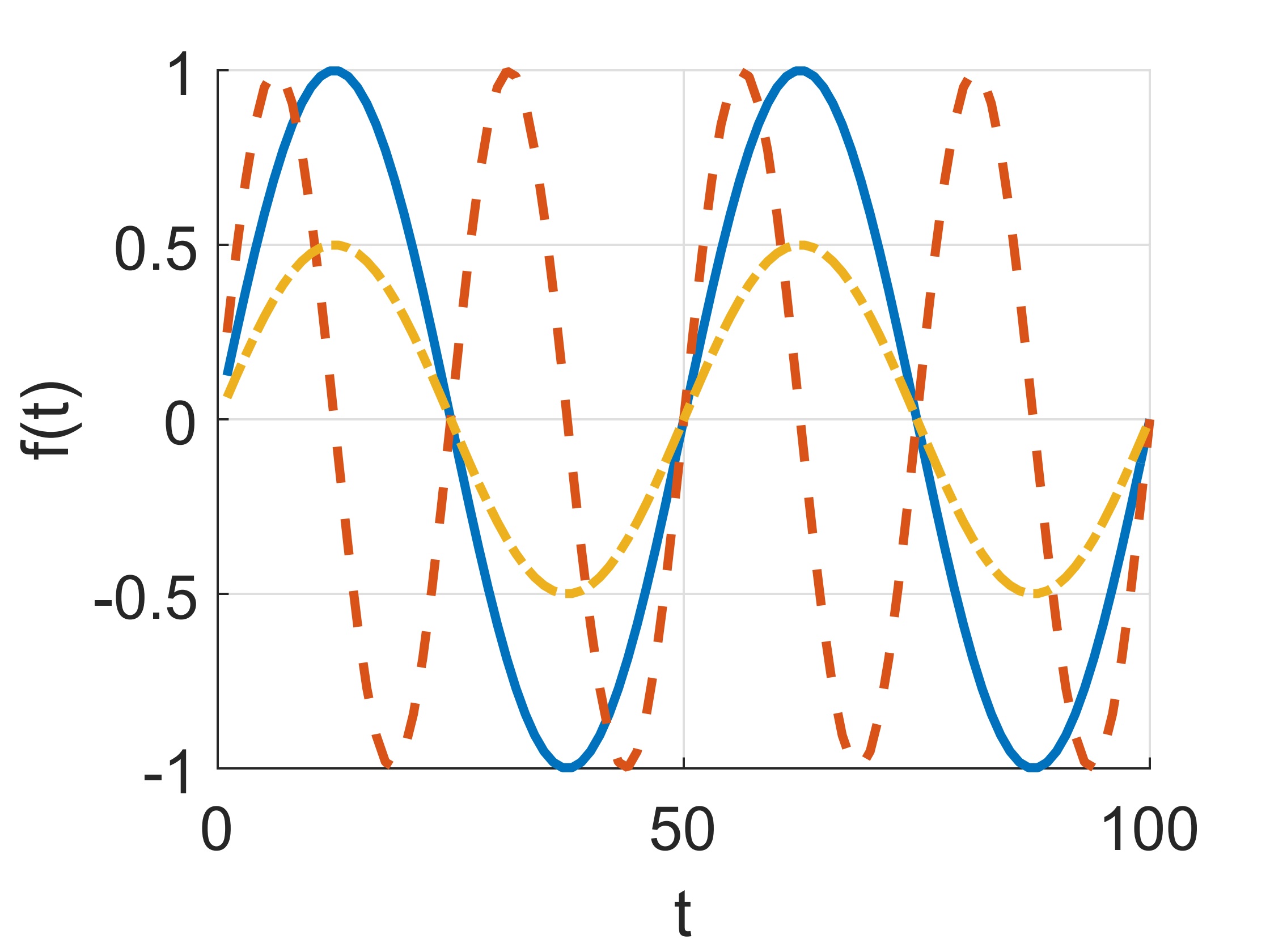}}
	\subfigure[Delay embedding]{\label{subfig:DE_comp_b}
		\includegraphics[width=.45\columnwidth, trim=5 5 20 15, clip]{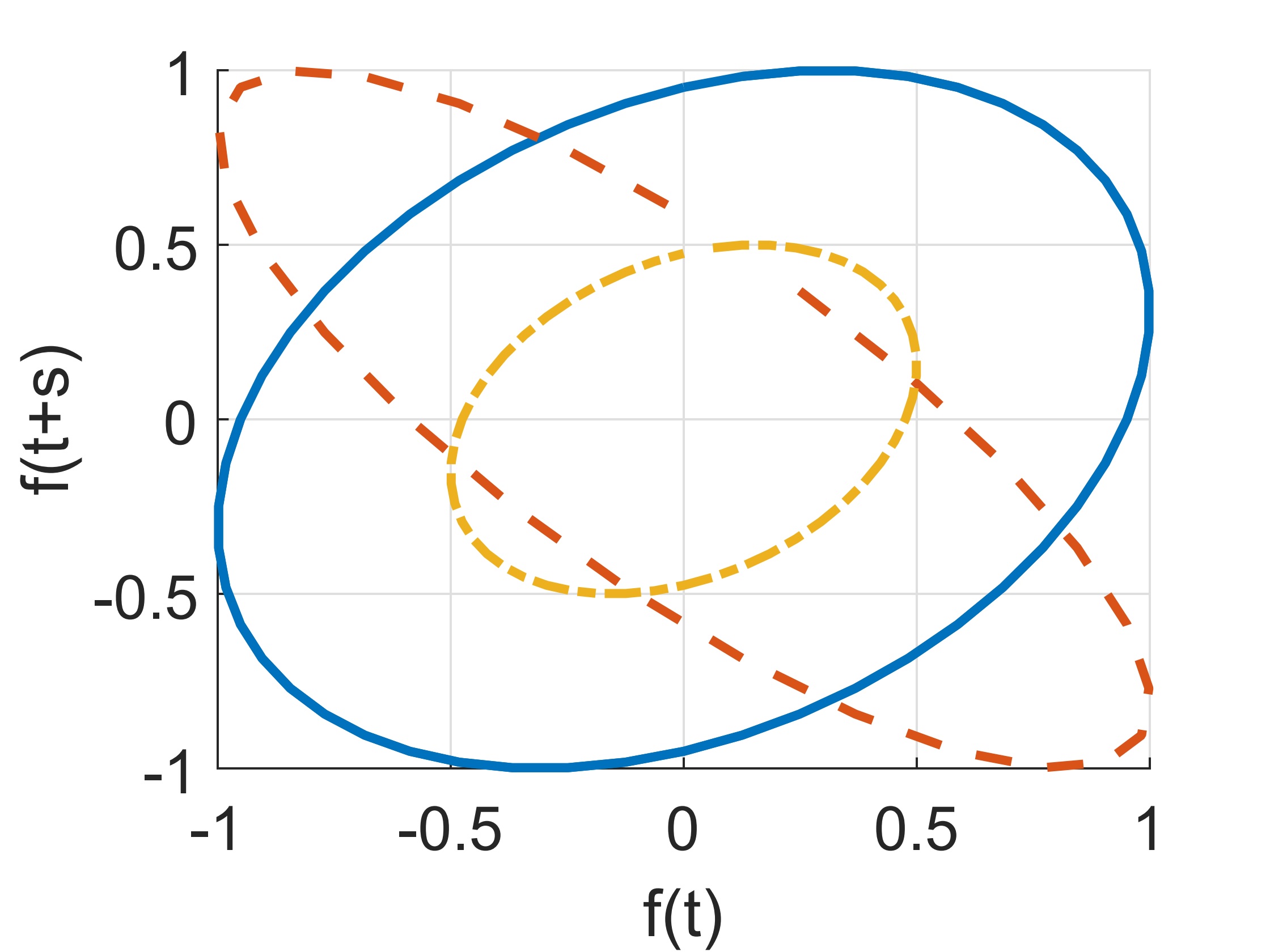}}
	\caption{Delay embedding on time series with different patterns. (a) Time series of different periods or amplitudes. (b) The corresponding results from delay embedding.}
	\label{fig:ED_comp}
\end{figure}

A toy example of delay embedding is shown in Fig.~\ref{fig:DE}, assuming a 1-D time series $y_t=f(t)$, $t\in\mathbb{Z}^+$, $d=2$, and $s=1$. For each delay embedding, only the adjacent two observations are involved, and the time series is transformed to the embedding space, where the raw 1-D time series becomes a recursive 2-D time series. According to \cite{perea2013sliding}, the trajectory of the recursive 2-D time series forms a pattern corresponding to the intrinsic characteristics of data in the time domain. Fig.~\ref{fig:ED_comp} illustrates that different patterns in the time series result in different trajectories in the embedding space. Intuitively, we can classify the time series through their trajectories, which is invariant to the phase and length changes of the time series.

\subsection{Derivative Delay Embedding}
Although delay embedding is robust to the length and phase changes of the time series, as shown above, it is sensitive to the shift of baseline. For example, the zero-drift effect will make the sensor output drift away although the external environment has not changed at all; or different types of sensors monitoring the same variable may yield results in different baseline.    
Moreover, the embedding space is a continuous space, so recording the exact position of the states/trajectory would consume large memory. 

\begin{figure}[h]
	\centering
	\includegraphics[width=1\columnwidth, trim=10 5 10 0, clip]{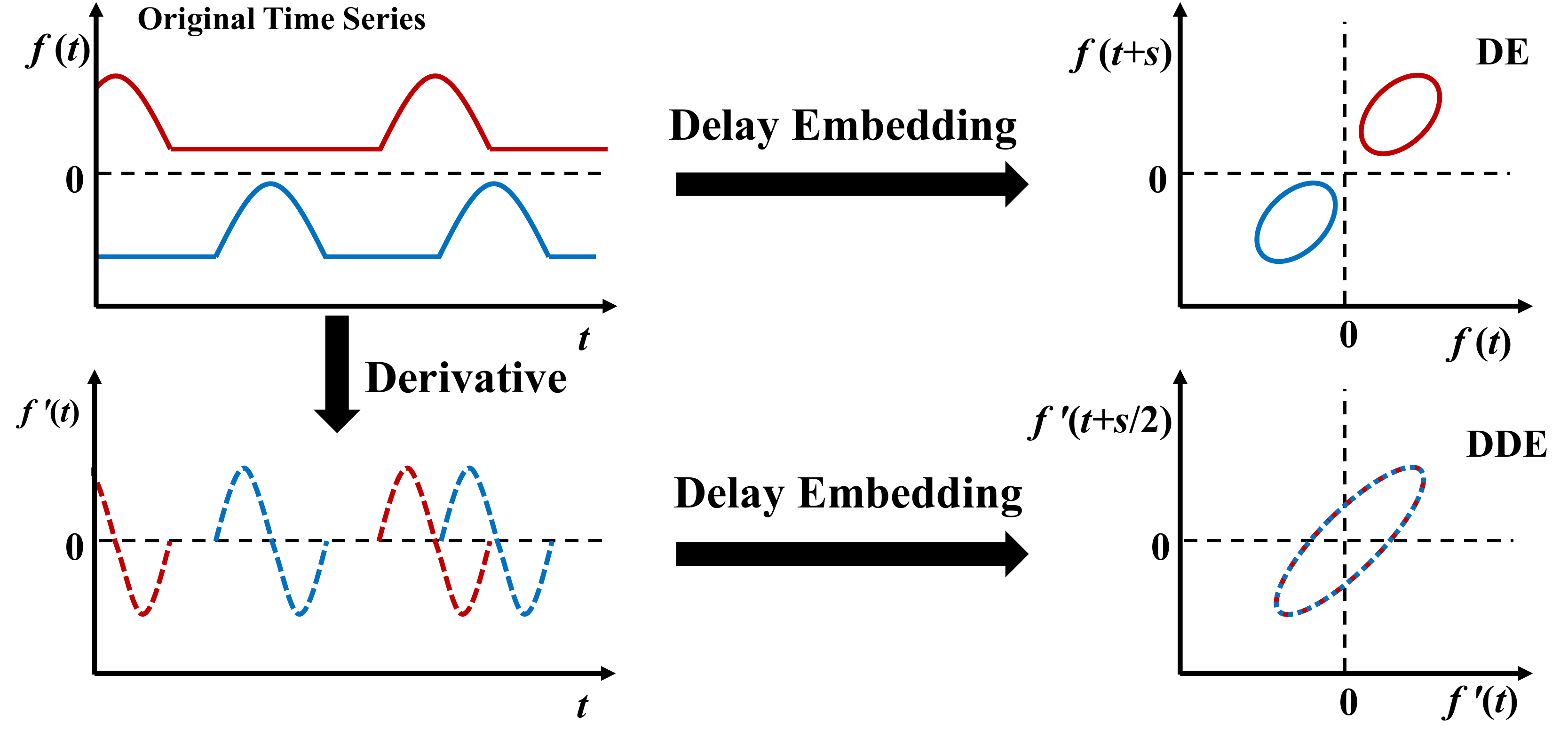}
	\caption{Comparison of DE and DDE. The time series (top left) in red and blue have the same pattern but different phase and baseline (misalignment). The delay embedding results in two trajectories with the same shape but different locations, while DDE (bottom right) generates exactly the same trajectory for both time series.}
	\label{fig:DDE}
\end{figure}

To tackle these two problems, we develop the derivative delay embedding (DDE) method, letting the observation function $\psi(x_t)=y'_t$ to offset the baseline shift, and then the embedding space is discretized into a grid to reduce the memory cost. The DDE of $y_t=f(t)$ at $t\in\mathbb{Z}^+$ is 
\begin{equation}
\Phi'(x_t)=G\left( y'_t, y'_{t+s},\cdots,y'_{t+(d-1)s} \right),
\label{eq:DDE}
\end{equation}
where $y'_t=\left(y_t-y_{t-\tau}\right)/\tau$, $\tau\in\mathbb{Z}^+$. $G(\cdot)$ approximates a state to the nearest grid cell in the discretized embedding space. For simplicity, $\tau$ is set to 1, thus $y'_t=y_t-y_{t-1}$. An illustrative comparison between the delay embedding and DDE is shown in Fig.~\ref{fig:DDE}, assuming $y_t=f(t)$.

Because the derivative intrinsically has a zero baseline, the DDE gains the invariance to the shift of baseline, enabling the complete relaxation of those common but unrealistic assumptions, i.e., the same length and well alignment. 

On the other hand, DDE generates recursive trajectories that occupy a limit region of the embedding space, the discretized embedding space further realizes an approximately constant footprint. In Fig.~\ref{fig:DE}, for example, the raw time series consists of 15 points. After delay embedding, there are only 8 points in the embedding space because of recursiveness. In practice, however, the time series is normally corrupted by noise which will prevent the trajectory from presenting such perfect recursiveness as shown in Fig.~\ref{fig:DDE}. 
\begin{figure}[h]
	\centering
	\subfigure[Time series]{\label{subfig:DDE_grid_a}
		\includegraphics[width=.36\columnwidth, trim=5 5 20 30, clip]{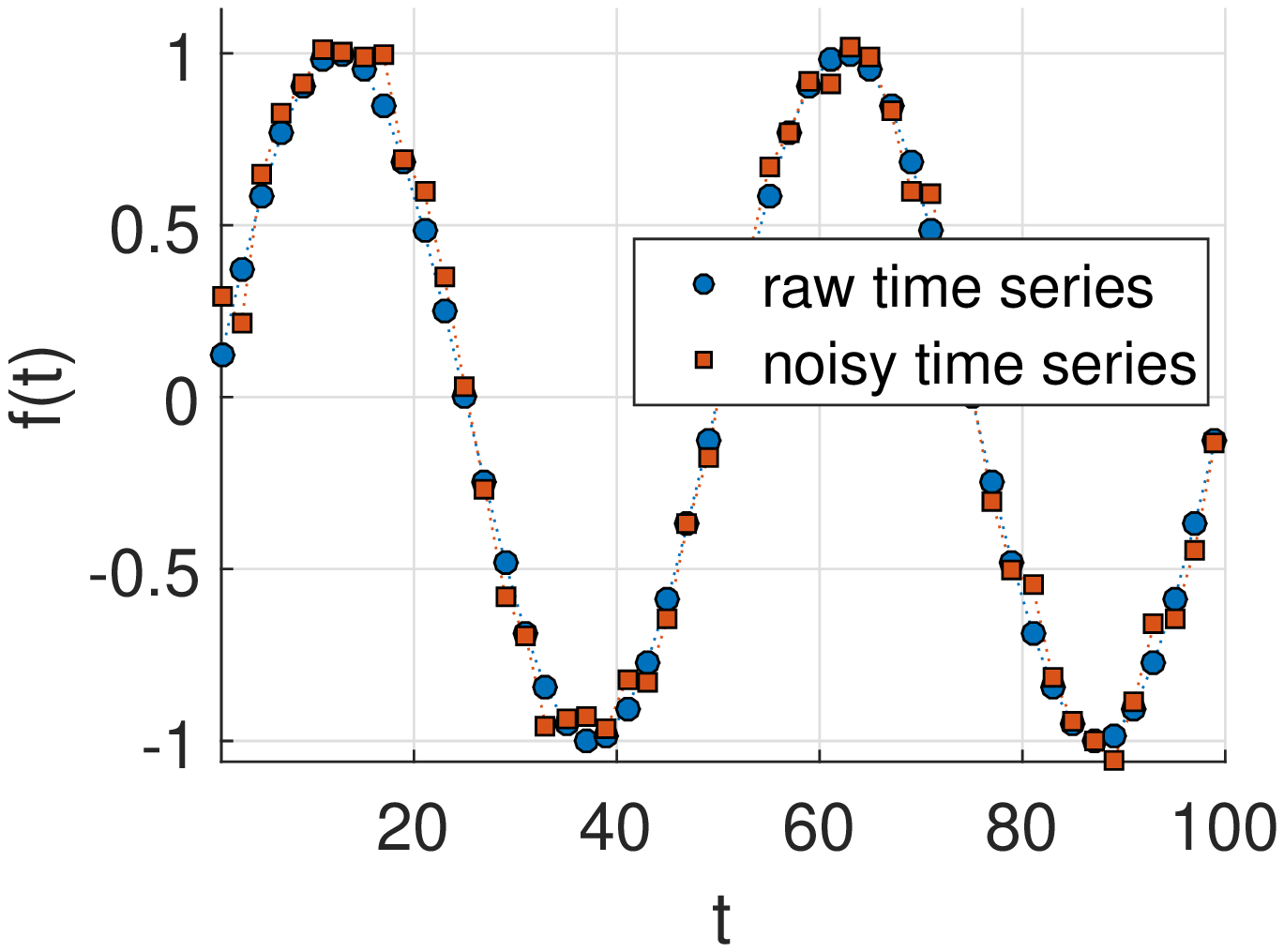}}
	\subfigure[Embedding]{\label{subfig:DDE_grid_b}
		\includegraphics[width=.27\columnwidth, trim=30 0 95 30, clip]{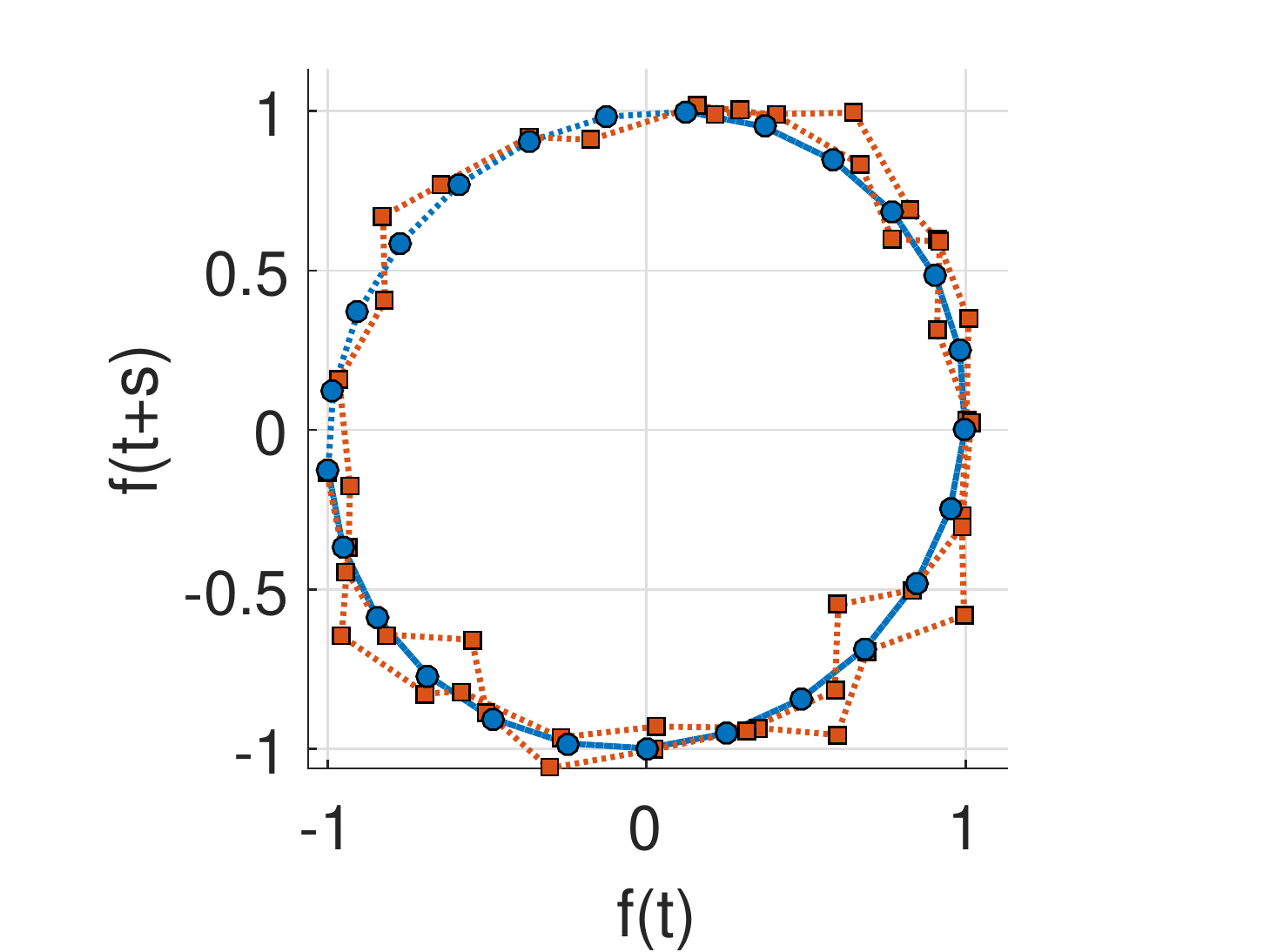}}
	\subfigure[Discretization]{\label{subfig:DDE_grid_c}
		\includegraphics[width=.3\columnwidth, trim=10 0 60 15, clip]{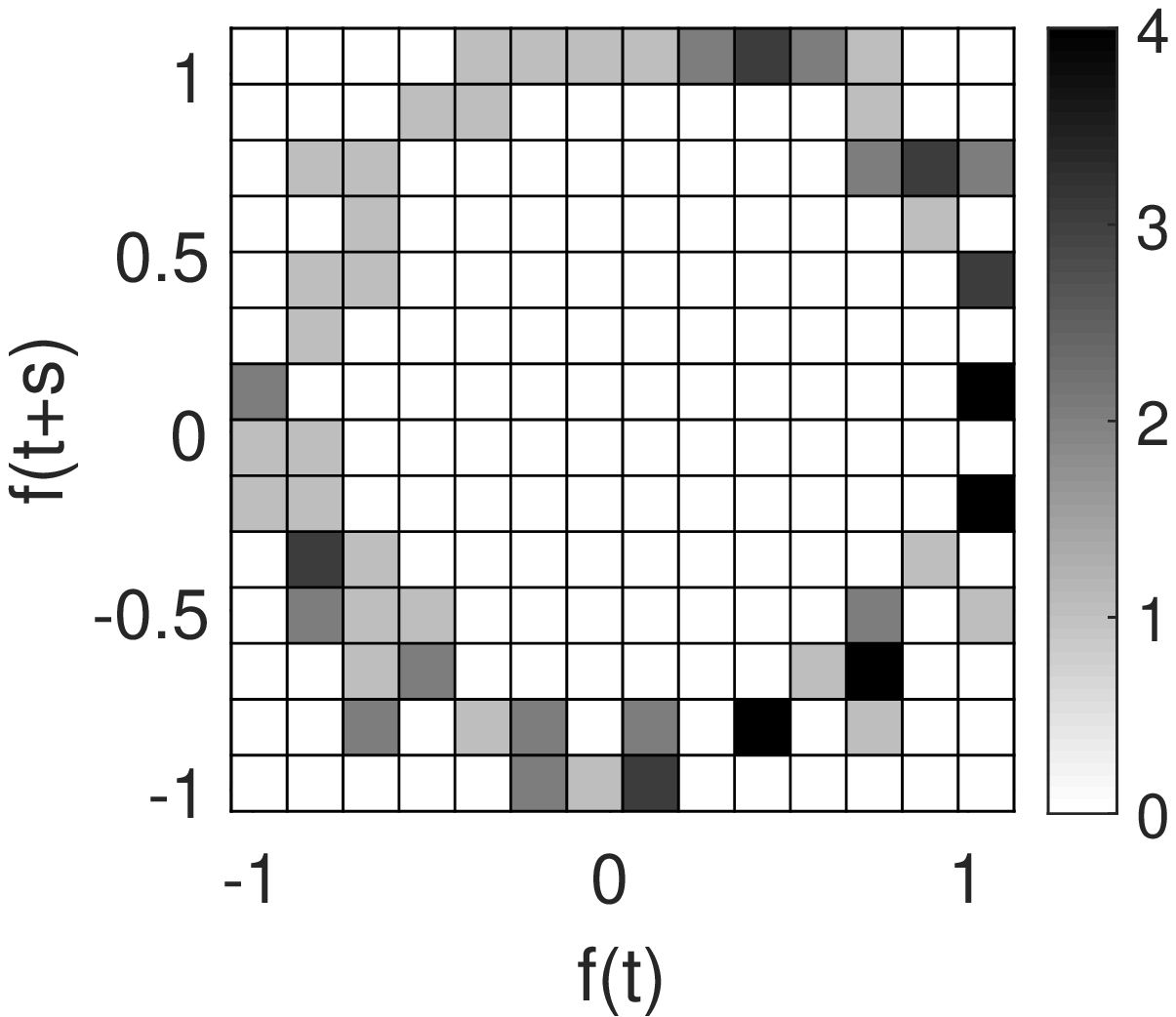}}
	\caption{A toy example of discretized embedding space. (a) The raw time series (blue circle) and noisy one (red square). (b) The corresponding trajectories in the embedding space. (c) The discretized embedding space. The color bar indicates the number of states falling into each grid cell. }
	\label{fig:DDE_grid}
\end{figure}
Therefore, the number of unrepeated states in the embedding space will be similar to the number of points in the raw time series, and then the memory consumption of storing all the states will not be less than that of recording the raw time series. 
In the discretized embedding space (Fig.~\ref{fig:DDE_grid}), however, if the size of grid cell is chosen appropriately, the deviated states caused by noise will fall into the same grid cell, which drastically reduces the memory cost and achieves denoising effect at the same time. In addition, any recursive trajectory can be represented by a finite number of cells in the discretized embedding space. 
Therefore, the discretization drastically reduces the memory consumption and potentially preserves a constant memory footprint although handling an infinite time series.
Intuitively, the memory cost is decided by the cell size --- a smaller cell size requires larger memory and vice versa. 
Section~\ref{subsec:paramter_selection} will discuss more details on the choice of parameters.

\subsection{Parameter Selection}  
\label{subsec:paramter_selection}
There are two parameters in delay embedding, i.e., the delay step $s$ and embedding dimension $d$.
From empirical study, $d$ and $s$ both significantly affect the performance of delay embedding in the aspect of classification accuracy, and they are application orientated, varying with different datasets. In DDE, we have an extra parameter --- the cell size of the discretized embedding space that decides the fidelity of representing the state trajectory. This section discusses the methods of selecting appropriate values for $s$, $d$, and the grid size.
  
\subsubsection{Delay Step ($s$)}
According to \cite{perc2005dynamics}, an effective method of obtaining the optimal $s$ is to minimize the mutual information~\cite{fraser1986independent} between $y_t$ and $y_{t+s}$. The idea is to ensure a large enough $s$ so that the information measured at $t+s$ is significantly different from that at time $t$. However, it needs to manually divide the observation into equally sized bins in order to compute the mutual information. 
\cite{perea2013sliding} provided a criterion to obtain the optimal $s$ based on periodic time series,
\begin{equation}
\centering
2\pi\times d\times s\times\frac{f}{f_s}\equiv 0\mod{\pi},
\label{eq:s}
\end{equation} 
where $f$ and $f_s$ denote the resonant and sampling frequency, respectively, of the time series. In practice, however, the time series is not necessarily periodic. 
Based on the ideas from \cite{perc2005dynamics,perea2013sliding}, we decide $s$ based on the dominant frequency of the raw time series. Instead of assigning the resonant frequency to $f$ as in Eq.~\ref{eq:s}, we adopt the dominant frequency --- the frequency with the maximum magnitude not counting the DC component in the frequency domain. 
To obtain an appropriate $s$, let $2\pi\times d\times s\times f/f_s=\pi$, which minimizes the information loss when transforming the time series to the embedding space because this is the case that yields the smallest $s$ from Eq.~\ref{eq:s}. At the same time, the minimum $s$ is bounded by $f_s/(2\times d\times f)$ to avoid large mutual information.

Practically, a times series is a sequence of points of length $N$. Applying Fast Fourier Transform (FFT)~\cite{brigham1988f}, we can obtain the dominant frequency $f=nf_s/N$, where $ n $ denotes the index of the maximum magnitude in the spectral space. Therefore, a more succinct formula of $s$ is
\begin{equation}
\centering
s=\frac{N}{2d\times n}.
\label{eq:s2}
\end{equation} 

\begin{figure}[h]
	\centering
	\subfigure[Time series]{\label{subfig:cmp_s_signal}
		\includegraphics[width=.6\columnwidth, trim=30 5 50 10, clip]{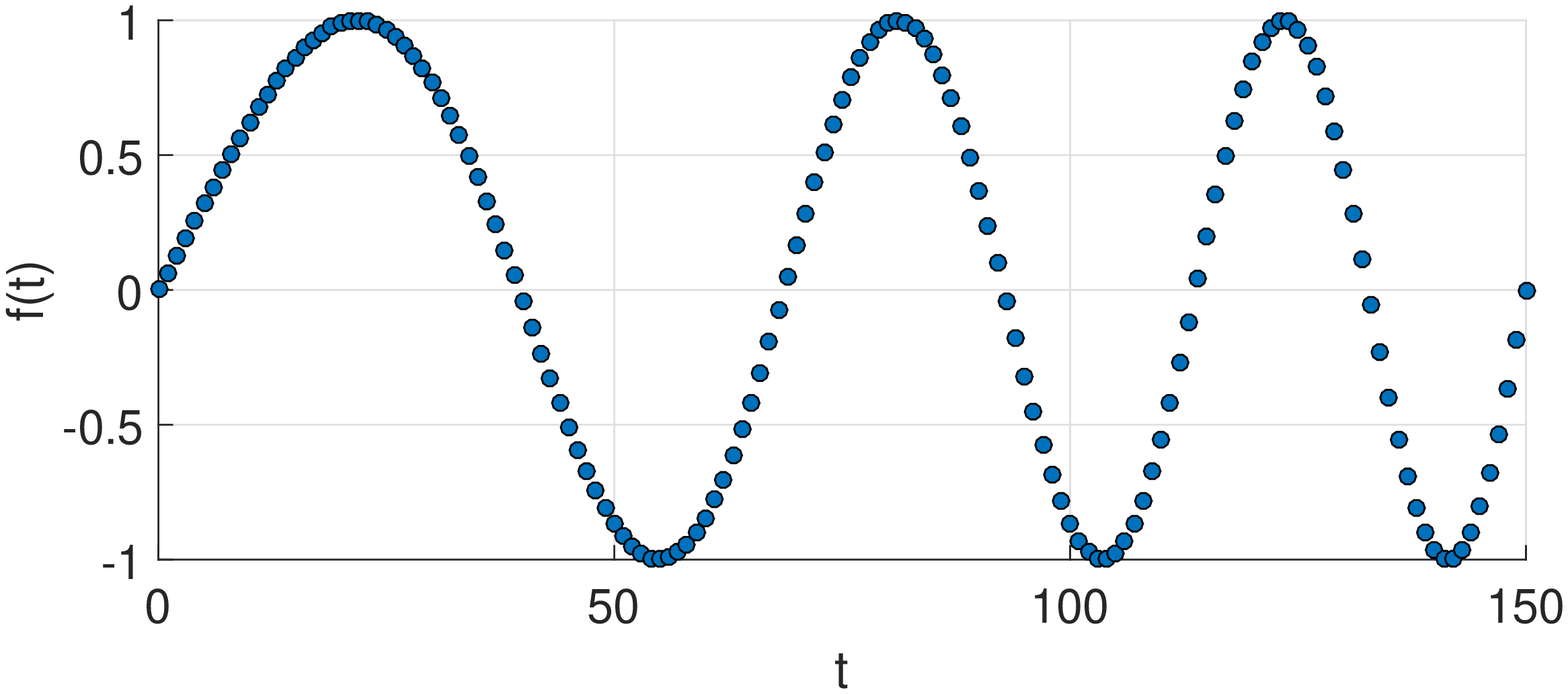}}
	\subfigure[FFT]{\label{subfig:cmp_s_fft}
		\includegraphics[width=.35\columnwidth, trim=5 0 40 10, clip]{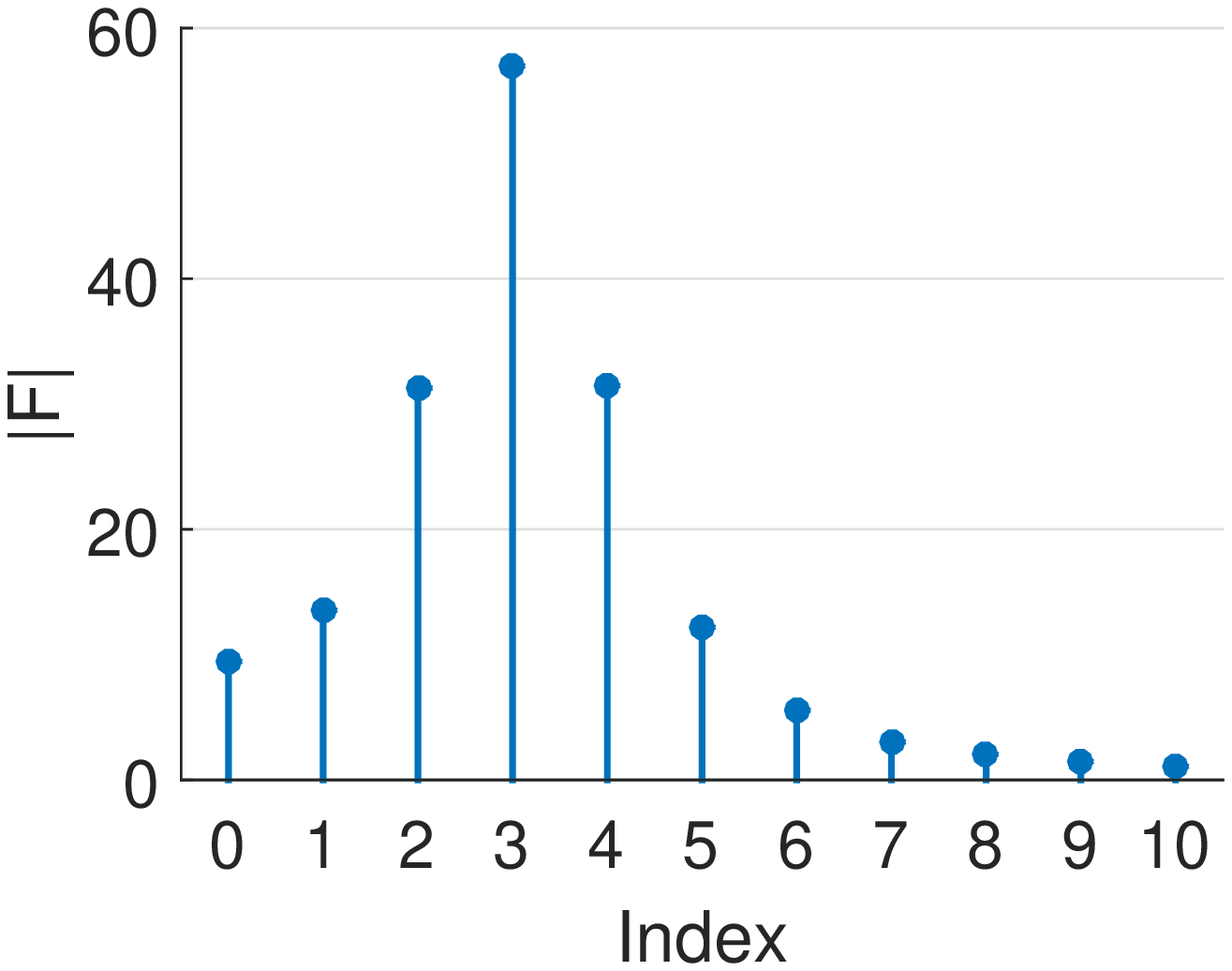}}\\
	\subfigure[$s=2$]{\label{subfig:cmp_s_de_2}
		\includegraphics[width=.3\columnwidth, trim=30 0 90 20, clip]{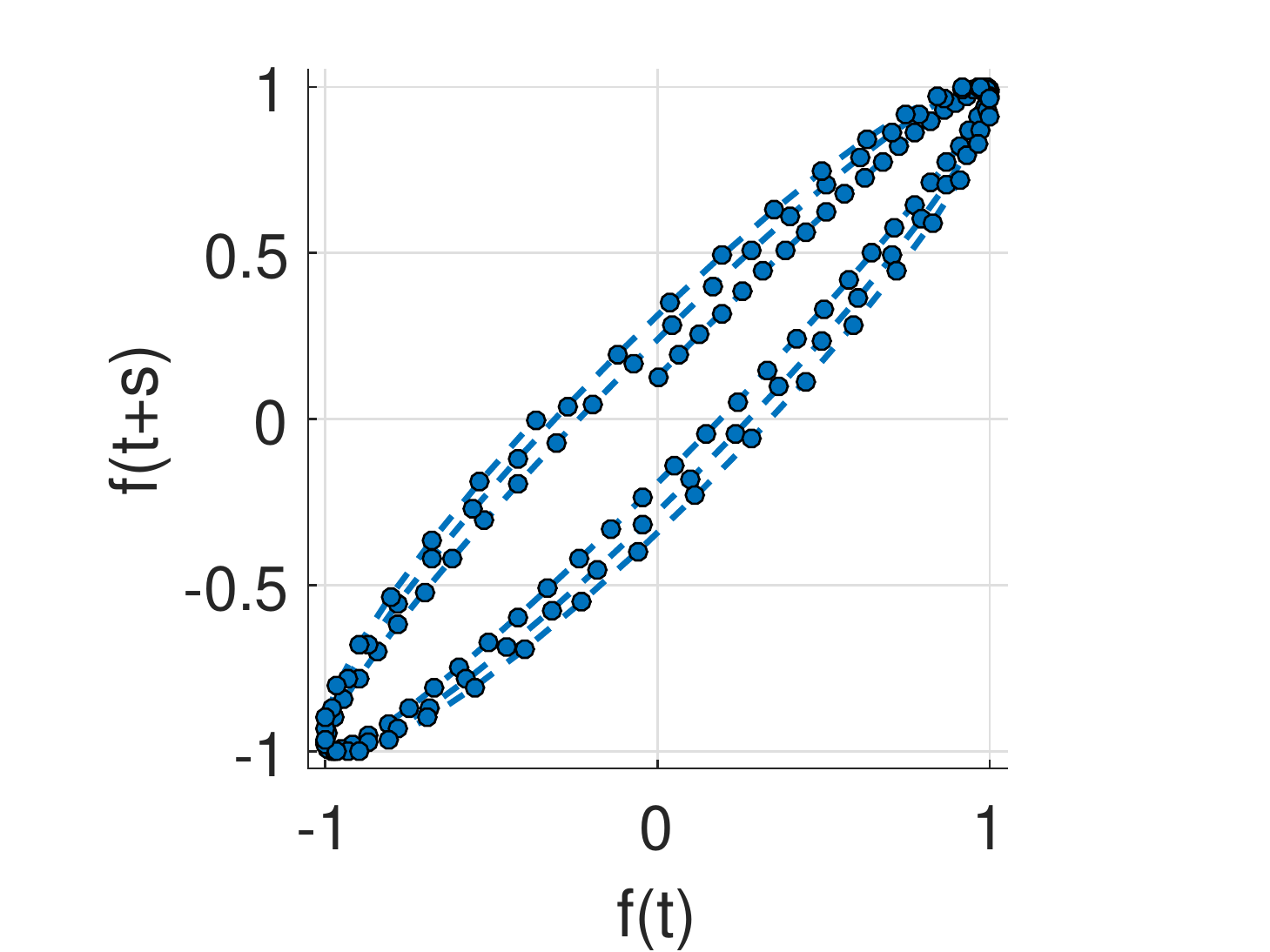}}
	\subfigure[$s=12$]{\label{subfig:cmp_s_de_12}
		\includegraphics[width=.3\columnwidth, trim=30 0 90 20, clip]{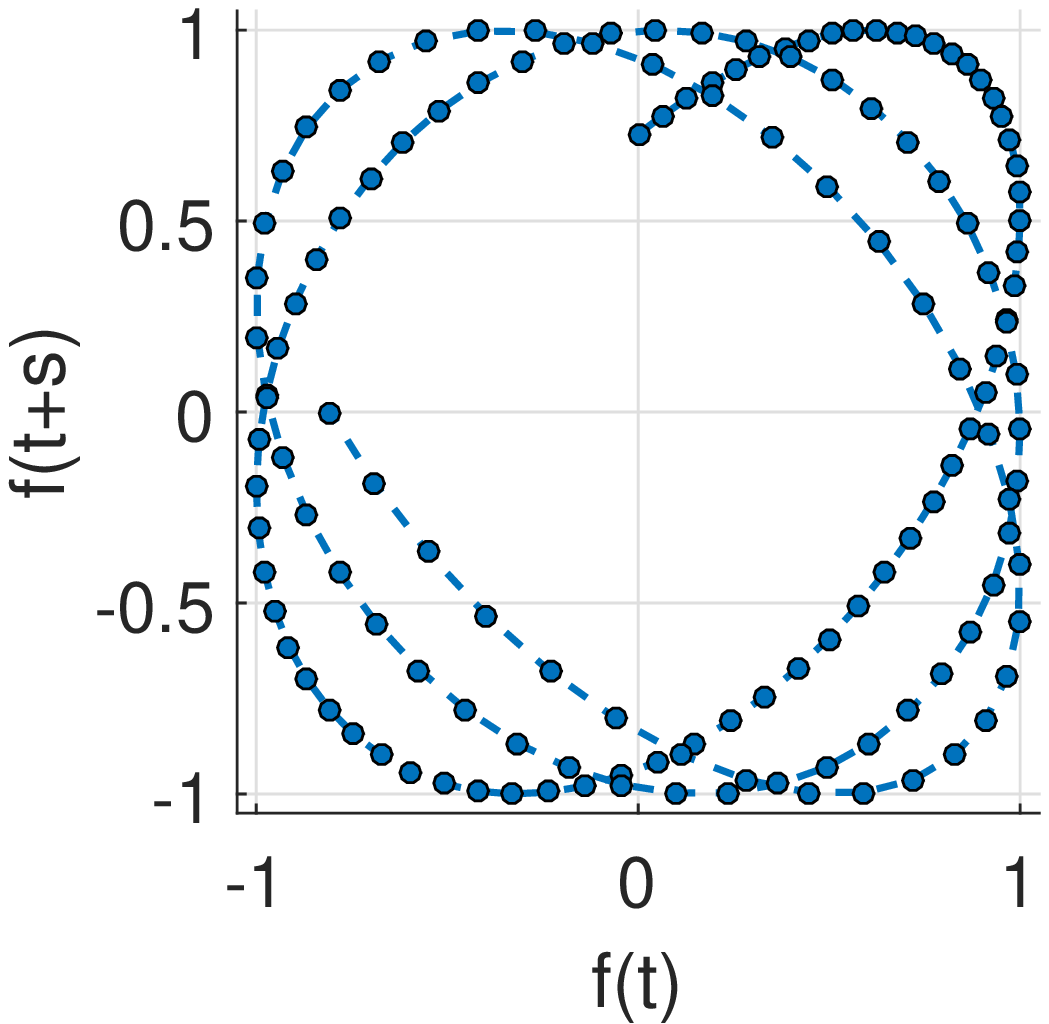}}
	\subfigure[$s=25$]{\label{subfig:cmp_s_de_25}
		\includegraphics[width=.3\columnwidth, trim=30 0 90 20, clip]{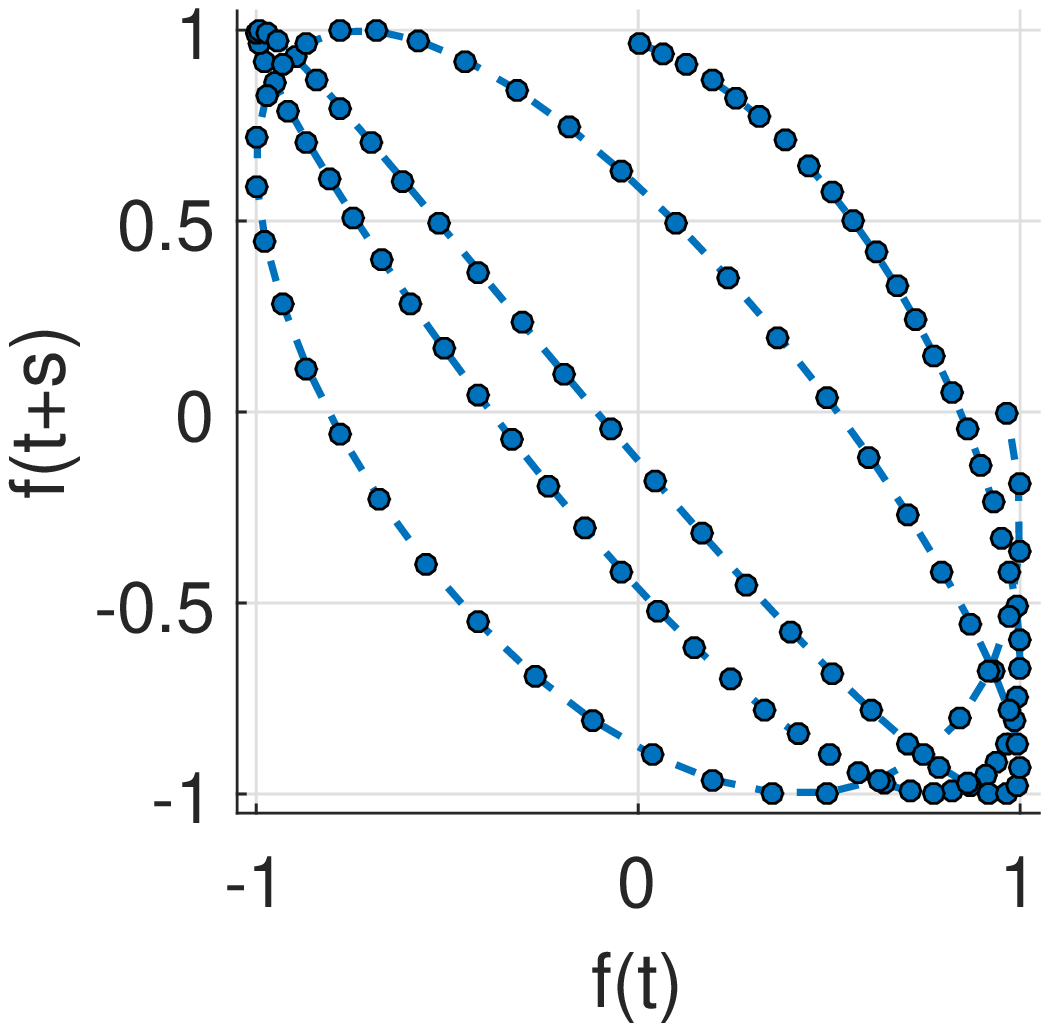}}
	\caption{Effect of delay step $s$ on delay embedding. (a) Time series with increased frequency. (b) The spectrum of the time series from FFT. (c), (d) and (e) The corresponding results from delay embedding with different step sizes.}
	\label{fig:cmp_s}
\end{figure}

Fig.~\ref{fig:cmp_s} illustrates the selection of $s$. Fig.~\ref{subfig:cmp_s_signal} shows a time series with the length of $N=151$ points. The index of the dominant frequency from FFT is $n=3$ as shown in Fig.~\ref{subfig:cmp_s_fft}. From Eq.~\ref{eq:s2}, an appropriate step size for $d=2$ is $s=151/(2\times 2\times 3)\approx 12.58$. Since $s$ must be an integer, we set $s=12$. Comparing Figs.~\ref{subfig:cmp_s_de_2}, \ref{subfig:cmp_s_de_12} and \ref{subfig:cmp_s_de_25}, the roundness of the trajectory is maximized when $s=12$. Either smaller or larger $s$ will result in more overlap (mutual information) which runs higher risk of misclassification.

\subsubsection{Embedding Dimension ($d$)}
To determine a proper embedding dimension $d$, we apply the false nearest neighbor method developed in \cite{kennel1992determining}. This method assumes that the states that are close in the embedding space have to stay sufficiently close during forward iteration. If a reconstructed state has a close neighbor that does not fulfill this criterion, it is marked as having a false nearest neighbor. The steps for finding the optimal $d$ are:
\begin{enumerate}
	\item Given a state $\Phi(x_i)$ in the $d$-dimensional embedding space, find a neighbor $\Phi(x_j)$ so that $\|\Phi(x_i)-\Phi(x_j)\|_2<\varepsilon$, where $\varepsilon$ is a small constant usually not larger than $1/10$ of the standard deviation of the time series.
	\item Based on the neighbors, compute the normalized distance $R_i$ between the $(m+1)$th embedding coordinate of state $\Phi(x_i)$ and $\Phi(x_j)$:
	\begin{equation}
		\centering
		R_i=\frac{\|y_{i+d\times s}-y_{j+d\times s}\|_2}{\|\Phi(x_i)-\Phi(x_j)\|_2}
		\label{eq:Ri}
	\end{equation}
	\item If $R_i$ is larger than a given threshold $R_{th}$, then $\Phi(x_i)$ is marked as having a false nearest neighbor.
	\item Apply Eq.~\ref{eq:Ri} for the whole time series and for various $m = 1, 2,\cdots$ until the	fraction of points for which $R_i > R_{th}$ is negligible. According to \cite{kennel1992determining}, $R_{th} = 10$ has proven to be a good choice for most data sets.
\end{enumerate}

\begin{figure}[h]
	\centering
	\includegraphics[width=.5\columnwidth]{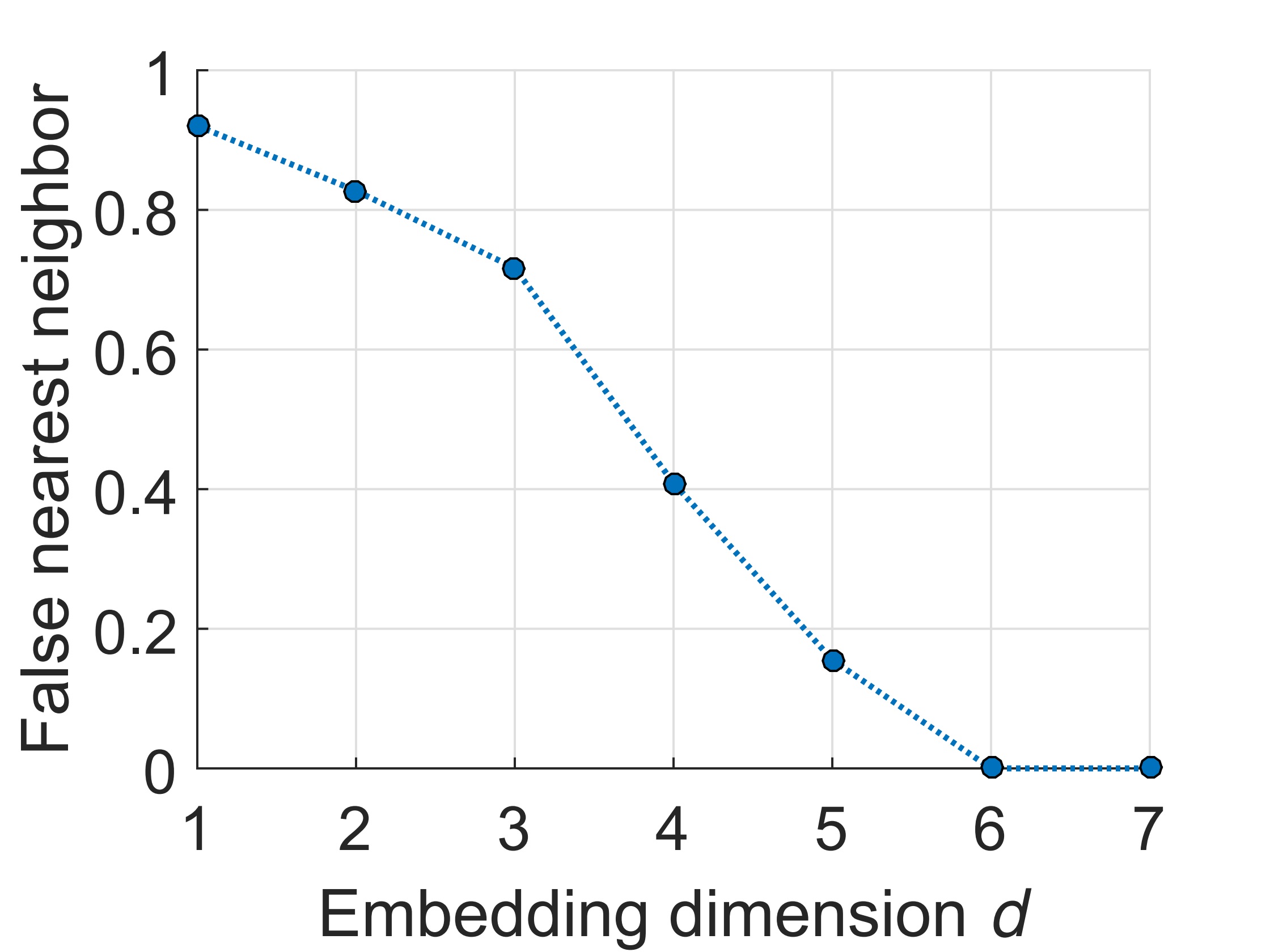}
	\caption{Selection of embedding dimension $d$. The raw time series is shown in Fig.~\ref{subfig:cmp_s_signal}, and $s=12$. The optimal embedding dimension is $d=6$ because the false nearest neighbor first achieves zero. In practice, $d=5$ is also an acceptable setting.}
	\label{fig:fnn}
\end{figure}

Applying the above method on the time series in Fig.~\ref{subfig:cmp_s_signal}, the process of finding the optimal $d$ is shown in Fig.~\ref{fig:fnn}.

In DDE, the above methods of finding appropriate $s$ and $d$ can be applied iteratively on the derivative of the raw time series. In practice, we cannot ensure the optimal parameter setting for any time series in classification tasks because the optimal setting always varies with classes, and we have to use a uniform setting for all classes to achieve fair data representation. Therefore, we randomly choose some training examples from each class and use the mean of the optimal settings from each class as the final values. The setting of $s$ only affects the classification accuracy, while $d$ also affects computational complexity. A larger $d$ does not necessarily improve the classification accuracy but certainly increases the burden on computation. To balance the accuracy and computational complexity, we prefer to select a smaller $d$.

\subsubsection{Size of Grid Cell}
The third parameter is the cell size of the discretized embedding space used in DDE. The cell size decides the fidelity of representing the trajectories, which in turn affects the classification accuracy. Generally, the accuracy increases as the cell size decreases, nevertheless, a too small cell size drastically increases the computational complexity and memory cost. 
Actually, when the cell size goes smaller and smaller, the overfitting problem starts to surface and the model ends up fitting the noisy data.
%
From our experiment, an appropriate cell size is $(y'_{\max}-y'_{\min})/50$, where $y'_{\max}$ and $y'_{\min}$ denote the maximum and minimum of the derivative time series, respectively. In other words, the trajectories are represented on a grid with approximately 50 bins on each dimension.

\section{Markov Geographic Model}
As discussed in DDE, the trajectory constructed from a time series preserves distinguishable and robust patterns. Therefore, we can model the trajectories of different classes during training. Then, the label of a testing time series could be decided by comparing with those learned trajectories. Many existing works related to delay embedding would model the trajectories by a group of differential functions, parametric models~\cite{lainscsek2015delay}, or topological features~\cite{perea2013sliding}, e.g., barcodes from persistent homology. However, they all perform in an offline manner, and it is difficult to find a parametric model that is suitable for all applications. We propose a non-parametric model MGM that could model the trajectories in an online manner. 

From Fig.~\ref{fig:DE}, we can see that the trajectory and geographical location of the states both carry significant information that distinguish one pattern from another. The trajectory can be modeled by the \textit{Markov process} --- the arrows in the embedding space of Fig.~\ref{fig:DE} denote transition of the states. 
However, the Markov process is sensitive to the probability of the starting state, e.g., if the starting state is an outlier, the probability of starting state will be small, thus the final probability of the whole trajectory will be small although the transition probability is large. Therefore, the \textit{geographic distribution} of the states is constructed instead which depicts the probability that a trajectory belongs to a class in a more global and robust manner. We refer to the proposed model as the Markov geographic model (MGM), which efficiently and effectively models both the geographic distribution of the states as well as their transition --- the two pieces of information that non-parametrically identify the deterministic function $\phi$ in Eq.~\ref{eq:evolve}. 

Specifically, the geographic distribution is represented by a probability map with the same size as the discretized embedding space. The state transitions are exhaustively recorded by an ``expandable list''. When a new transition appears, it is appended to the end of the list if it has not occurred in the past, otherwise, it is accumulated to the existing transition.

Assuming a $ d $-D discretized embedding space, and each grid cell is associated with an accumulator, counting the number of states falling into the cell during training. Then, the geographic distribution of the states can be obtained by
\begin{equation}
\label{eq:CPM} 
P(x_t) = \frac{\log \left(|\Phi'(x_t)|+1\right)}{\sum_{i}\log \left(|\Phi'(x_i)|+1\right)},
\end{equation}
where $P(x_t)$ is the probability of the state $x_t$ belonging to the training trajectories of a given MGM, $|\Phi'(x_t)|$ returns the number of states falling into the grid cell of coordinate $\Phi'(x_t)$ in the discretized embedding space. Because the derivative of a time series would result in many zero-crossing points, significantly increasing the number of states falling around the origin ($[0]^d$) of the embedding space, we apply the logarithm to suppress these large counts.    

Similarly, the transition probability from a state to another can be expressed as
\begin{equation}
\label{eq:PM} 
P(x_t|x_{t-1})=\frac{|\Phi'(x_{t});\Phi'(x_{t-1})|}{\sum_{i}|\Phi'(x_{i});\Phi'(x_{t-1})|},
\end{equation} 
where $|\Phi'(x_{t});\Phi'(x_{t-1})|$ returns the number of transitions from $x_{t-1}$ to $x_t$ during modeling, and $x_i$ denotes the $i$th possible state transiting from $x_{t-1}$. 

Based on the Markov process and state distribution, the similarity between a testing trajectory and the MGM of a given class is defined by
\begin{equation}
\label{eq:MGM}
\begin{split}
S_{\text{MGM}}(X) & = \sum_{j=1}^{t}P(x_j)\prod_{i=2}^{t}P(x_i|x_{i-1})\\
~ & = S_{\text{G}}(X)\times S_{\text{M}}(X) 
\end{split},
\end{equation}
where $S_{\text{MGM}}(\cdot)$ is the similarity between the testing trajectory $X=[x_1,x_2,\cdots,x_t]$ and the MGM of the given class. $S_{\text{M}}(\cdot)$ and $S_{\text{G}}(\cdot)$ estimate the similarity in aspects of state transition and distribution, respectively. Compared to the original Markov process, we use the global probability of the whole trajectory as the starting probability instead of the single state probability to make it more robust to noise and outliers. 

\section{Online Modeling}
\label{subsec:online_modeling}
\begin{figure}[h]
	\centering
	\includegraphics[width=1\columnwidth]{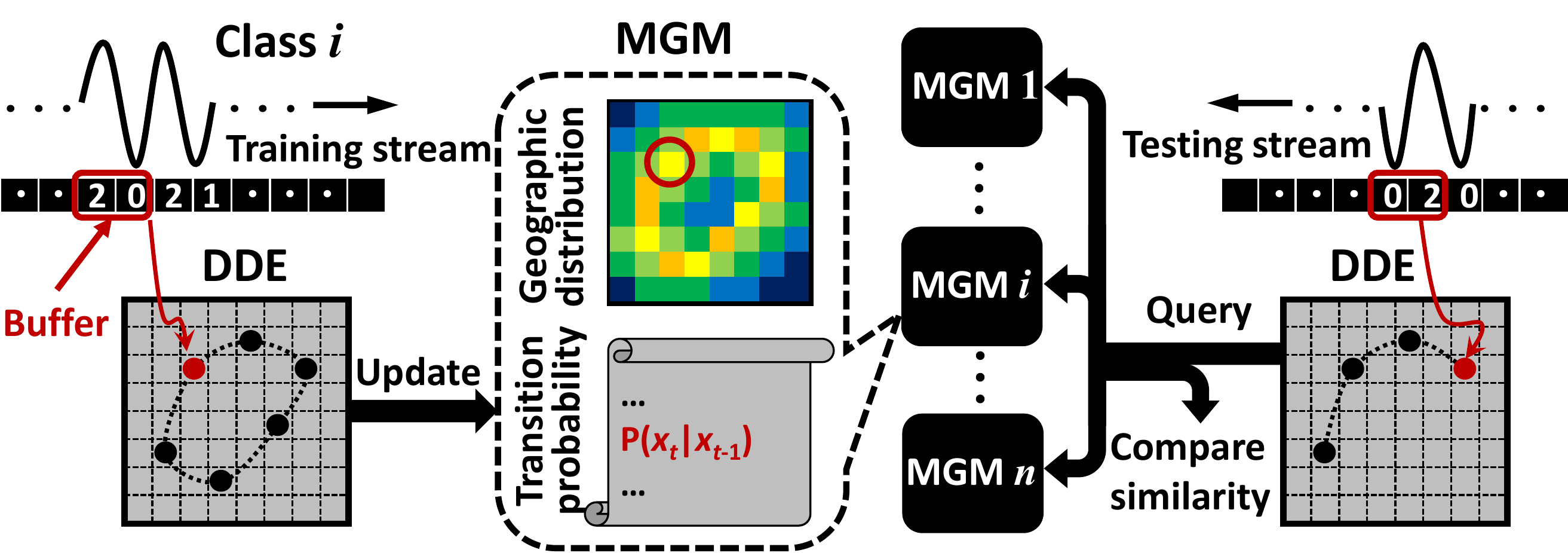}
	\caption{Flow of online modeling and classification based on DDE-MGM. The two stages can be performed in parallel.}
	\label{fig:flow}
\end{figure}

This section elaborates on how the proposed DDE-MGM  models and classifies the time series both in an online manner. Fig.~\ref{fig:flow} shows the flow of the DDE-MGM scheme, assuming $d=2$ and $s=1$ for simplicity.
During online modeling, a training time series streams through a buffer of size $(d-1)s+1=2$. When a new data point arrives, the oldest one in the buffer will be removed, and only the data points in the buffer are applied to DDE to construct a state in the discretized embedding space. Location of the state and its transition from its previous state are updated in the corresponding MGM. 

\subsection{Representing MGM}
In a discretized embedding space of $50\times 50$ grid, for example, the geographic distribution require $50^2$ bytes to record the counts ($|\Phi'(x_t)|$ in Eq.~\ref{eq:CPM}) of states falling into each cell. For the transition probability, a common way to record all possible transitions is to construct a $(50\times 50)^2$ matrix, which is huge and a waste of memory. Since such a matrix is normally sparse, we use a ``list'' to accumulate only those active transitions ($|\Phi'(x_{t});\Phi'(x_{t-1})|$ in Eq.~\ref{eq:PM}). 
Note that each class maintains a separate MGM. The modeling procedure transforms the time series and updates MGMs in real time without any preprocessing or making any assumptions on the time series. 
During the classification, a testing stream is also transformed to the embedding space, where Eq.~\ref{eq:MGM} can be applied to incrementally calculate the similarity between the testing stream and each class:
\begin{align}
	\centering
	\label{eq:update}
	& S_{\text{G}}^t=S_{\text{G}}^{t-1}+P(x_t),\\
	& S_{\text{M}}^t=S_{\text{M}}^{t-1}\times P(x_t|x_{t-1}),\\
	& S_{\text{MGM}}^t=S_{\text{G}}^t\times S_{\text{M}}^{t}.
\end{align}

\begin{algorithm}[t]
	\centering
	\caption{DDE-MGM online modeling/classification}
	\begin{algorithmic}[0]
		\STATE \textbf{Initialization} delay step $s$, embedding dimension $d$, and cell size of discretized embedding space
		\STATE \textbf{\hspace{10mm}*** Online Modeling Thread ***}
		\STATE \textbf{Input} a training stream $f(t)$
		\FOR {each time point $t$}
		\STATE obtain label index $i$
		\STATE $\Phi'(x_t)=G\left( f'(t), f'(t+s),\cdots,f'(t+(d-1)s) \right)$ 
		\STATE $|\Phi'(x_t)|_i=|\Phi'(x_t)|_i+1$
		\STATE $|\Phi'(x_t);\Phi'(x_t-1)|_i=|\Phi'(x_t);\Phi'(x_t-1)|_i+1$ 
		\STATE update $|\Phi'(x_t)|_i$ and $|\Phi'(x_t);\Phi'(x_t-1)|_i$ to $\text{MGM}_i$
		\ENDFOR
		\STATE \textbf{\hspace{10mm}*** Online Classification Thread ***}
		\STATE \textbf{Input} a testing stream $g(t)$
		\STATE \textbf{Output} label of $g(t)$ 
		\FOR {each time point $t$} 		
		\STATE $\Phi'(x_t)=G\left( g'(t), g'(t+s),\cdots,g'(t+(d-1)s) \right)$ 
		\FOR {each class $ i $}
		\STATE query $|\Phi'(x_t)|_i$ and $|\Phi'(x_t);\Phi'(x_t-1)|_i$ from $\text{MGM}_i$
		\STATE compute $P(x_t)$ and $P(x_t|x_{t-1})$ by Eqs.~\ref{eq:CPM} and \ref{eq:PM}
		\STATE $S_{\text{G}_i}^t=S_{\text{G}_i}^{t-1}+P(x_t)$
		\STATE $S_{\text{M}_i}^t=S_{\text{M}_i}^{t-1}\times P(x_t|x_{t-1})$
		\STATE $S_{\text{MGM}_i}^t=S_{\text{G}_i}^t\times S_{\text{M}_i}^{t}$
		\ENDFOR
		\IF {output required}
		\RETURN $\arg\underset{i}{\max}\{S^t_{\text{MGM}_i}\}_{i=1,2,\cdots}$ 
		\ENDIF
		\ENDFOR		
	\end{algorithmic}
	\label{alg:DDE-MGM}
\end{algorithm}

\subsection{Neighborhood Matching}
\begin{figure}[H]
	\centering
	\includegraphics[width=.6\columnwidth]{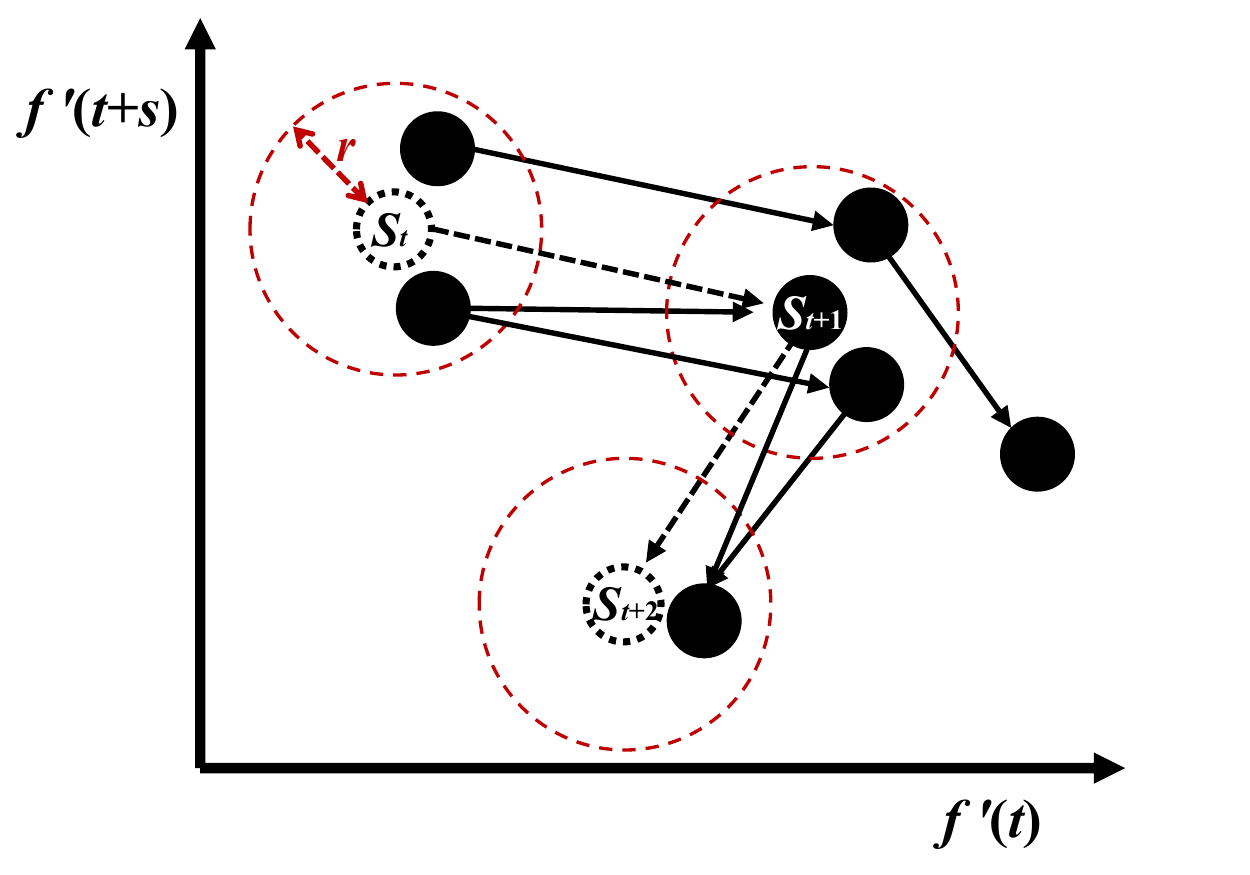}
	\caption{A toy example of the neighborhood matching. The black solid circles and arrows are learned states and transitions, respectively. The state marked with $S_t$, $S_{t+1}$, and $S_{t+2}$ denote the testing states, and the dashed arrows show the transitions. The dotted red circles indicate the neighbor region of the testing states within radius $r$.}
	\label{fig:neighbor}
\end{figure}

In practice, a testing trajectory cannot perfectly match an MGM. Therefore, we further propose the \textit{neighborhood matching} strategy to improve the robustness. The improved $S_{\text{M}}(X)$ based on neighborhood matching is defined as
\begin{equation}
\label{eq:score} 
S_{\text{M}}(X)= \prod_{i=2}^{t}\frac{\sum_{\alpha\in N_r(\Phi'(x_{i})),\;\beta\in N_r(\Phi'(x_{i-1}))}|\alpha;\beta|}{\sum_k\sum_{\gamma\in N_r(\Phi'(x_{i-1}))}|\Phi'(x_k);\gamma|},
\end{equation}
where $N_r(\Phi'(x_i))$ denotes the set of neighbors within radius $r$ around $\Phi'(x_i)$, and $k$ walks all possible states learned by the MGM. Usually, radius $r$ is set to be the cell size, which means the neighbors are searched from a $3 \times 3$ window.  The improved $S_{\text{M}}(X)$ forms clusters centered at the testing states, becoming an estimate of cluster-wise transition probability, which effectively increases the robustness to noise and intra-class variation. Fig.~\ref{fig:neighbor} illustrates the neighbor matching.

The DDE-MGM scheme is summarized in Algorithm~\ref{alg:DDE-MGM}, where the accumulators, i.e., $|\Phi'(x_t)|$ and $|\Phi'(x_t);\Phi'(x_t-1)|$, are simultaneously updated and queried by the modeling and classification threads. After the initial training stage, the classification thread can be performed in parallel with the modeling without having to wait until the end of the training stream. 

\section{Experimental Evaluation}
The proposed DDE-MGM is evaluated on three datasets --- UCI character trajectories~\cite{Lichman:2013}, MSR Action3D~\cite{wang2012mining}, and PAMAP~\cite{reiss2011towards} outdoor activities. To illustrate the low computational complexity and superior classification performance, DDE-MGM is compared to HMM~\cite{lv2006recognition}, SAX~\cite{lin2003symbolic} and 1NN-DTW~\cite{xi2006fast}, which are considered the best algorithms for time series classification. In addition, we also compare with some state-of-the-art online algorithms, 
RBP~\cite{cavallanti2007tracking},
Projectron~\cite{orabona2008projectron},
BPAS~\cite{wang2010online},  
BOGD~\cite{zhao2012fast}, and
NOGD~\cite{lularge2015}, 
to verify the online performance of DDE-MGM. Besides classification accuracy, the run time is also concerned to evaluate computational complexity. 

\subsection{Datasets}

	The UCI Character Trajectory dataset~\cite{Lichman:2013} consists of 2858 character samples of 20 classes. Three dimensions are kept --- x, y, and pen tip force. The data was normalized and shifted so that their velocity profiles best match the mean of the set. This dataset serves as the baseline to compare different algorithms because its samples are well aligned, truncated to the similar length and normalized to the same baseline. 
	
	The MSR Action3D dataset~\cite{wang2012mining} is the most popular dataset used by most action recognition related literature. It consists of 567 action samples of 20 classes performed by 10 subjects. This dataset is of random data length without careful alignment. Each action sample is presented by a sequence of skeletons with 20 joints in the 3-D spatial space. We consider such a sequence as a 60-D time series because each skeleton can be treated as a point of 20 (joints)~$\times$~3 (3-D space) dimensions. This dataset is to highlight the advantage of DDE-MGM in robustness to length variation and misalignment of the time series.
	
	The PAMAP outdoor activities dataset~\cite{reiss2011towards} was collected from wearable sensors on subjects' hand, chest, and shoe when performing physical activities --- walking very slow, normal walking, Nordic walking, running, cycling and rope jumping. The samples in this dataset last tens of minutes and do not have fixed length. Only the 3-D acceleration data on hand is used in the experiment, which is sufficient to warrant a competitive classification accuracy. 
	Because the samples is long in time and repetitive in patterns, e.g., walking for tens of minutes, this dataset is adopted to mainly examine the efficiency of DDE-MGM in aspects of run time and memory cost.

\subsection{Experimental Setup}
In DDE-MGM, there are four parameters in total---$s$, $d$ and grid size for DDE (Eq.~\ref{eq:DDE}), and $r$ for the neighborhood-matching-based similarity function (Eqs.~\ref{eq:MGM} and \ref{eq:score}). The parameters $s$ and $d$ can be obtained by applying Eqs.~\ref{eq:s2} and \ref{eq:Ri} (false nearest neighbor) on randomly selected training samples, and then choosing the averaged settings. Through extensive empirical studies, it is appropriate to divide each dimension of the embedding space into a roughly 50 bins. 
The size of one interval is set to be the cell size. 
The neighbor size $r$ is set to be the cell size.

In the experiment, two groups of algorithms are cited to compare with the propose DDE-MGM: 1) offline algorithms that can achieve the state-of-the-art classification accuracy but time-consuming or assuming the input data are of the same length and well aligned, and 2) online algorithms that learn models efficiently in an online fashion --- the model is updated and applied to testing alternately for each sample in the dataset. When a sample arrives, for example, the model is first applied to classify this sample, and then the sample is used to update the corresponding model. 

All algorithms are run with Matlab on a laptop with Intel i7 dual-core 2.4GHz CPU. Therefore, we can achieve a fair comparison on the run time. In the offline comparison, the classification accuracy is obtained by leave-50\%-out cross validation. In the online comparison, training and testing are performed alternatively on each sample of the dataset.

\subsection{Classification Performance}
In the comparison of classification performance, both accuracy and run time are considered. Parameter settings based on section~\ref{subsec:paramter_selection} for each dataset is listed in Table~\ref{tab:param}, where the cell size and neighbor size $r$ are not included because they can be considered as constant regardless of the different datasets. Compared to the MSR and PAMAP datasets, the dominant frequency of data samples in the UCI dataset is much smaller, so it is assigned a larger $s$ and $d$ to decrease the mutual information between the reconstructed states. 
\begin{table}[h]
	\centering
	\caption{Parameter setting for each dataset}
	\begin{tabular}{c|c|c}
		\hline
		Dataset & Delay step $s$ &  Embedding dimension $d$ \\
		\hline
		UCI & 8 & 5 \\
		MSR & 3 & 2 \\
		PAMAP & 5 & 2\\
		\hline
	\end{tabular}
	\label{tab:param}
\end{table}

\textbf{The UCI dataset:} 
Table~\ref{tab:UCI_cmp} compares the performance of DDE-MGM with both offline (upper block) and online (lower block) algorithms. The notation ``O/R'' is short for Online modeling/Random data length and alignment. ``+'' and ``-'' denote whether the algorithm is able to achieve O/R or not. The ``Time'' column shows the total run time of training and testing. From the aspect of run time, DDE-MGM cannot beat most online methods. For accuracy, however, DDE-MGM is superior to the state-of-the-art in both off- and on-line categories.  Note that the run time of DDE-MGM in the online testing is longer than that in the offline testing because the offline testing performs training and testing each on half of the dataset, while the online testing trains and tests alternatively on the whole dataset. Although DDE-MGM takes longer time on the whole dataset (2858 samples) in the online testing, it can still achieve real-time performance on any single sample. In addition, DDE-MGM realizes O/R in both training and testing.
\begin{table}[h]
	\centering
	\caption{Comparison on the UCI dataset}
	\begin{tabular}{c|c|c|c}
		\hline\hline
		Algorithm & Accu. (\%) & Time (sec) & O / R\\
		\hline
		1NN-DTW & 91.37 & 3.9$\times 10^4$ & - / +\\
		SAX & 89.96 & 128.85  &   - / -\\
		HMM & 57.89 & 7.4$\times 10^3$  &   - / -\\
		DDE-MGM & \textbf{92.07} & \textbf{34.21} & + / + \\
		\hline
		RBP & 92.62 & 9.44 & + / - \\
		Projectron & 92.62 & 110.26 & + / -\\
		BPAS & 94.68 & 22.81 & + / -\\
		BOGD & 90.02 & 15.24 & + / -\\
		NOGD & 91.65 & \textbf{9.04} & + / -\\
		DDE-MGM & \textbf{95.45} & 63.92 & + / + \\
		\hline\hline
	\end{tabular}
	\label{tab:UCI_cmp}
\end{table}

\begin{figure}[h]
	\centering
	\subfigure{\includegraphics[width=.48\columnwidth]{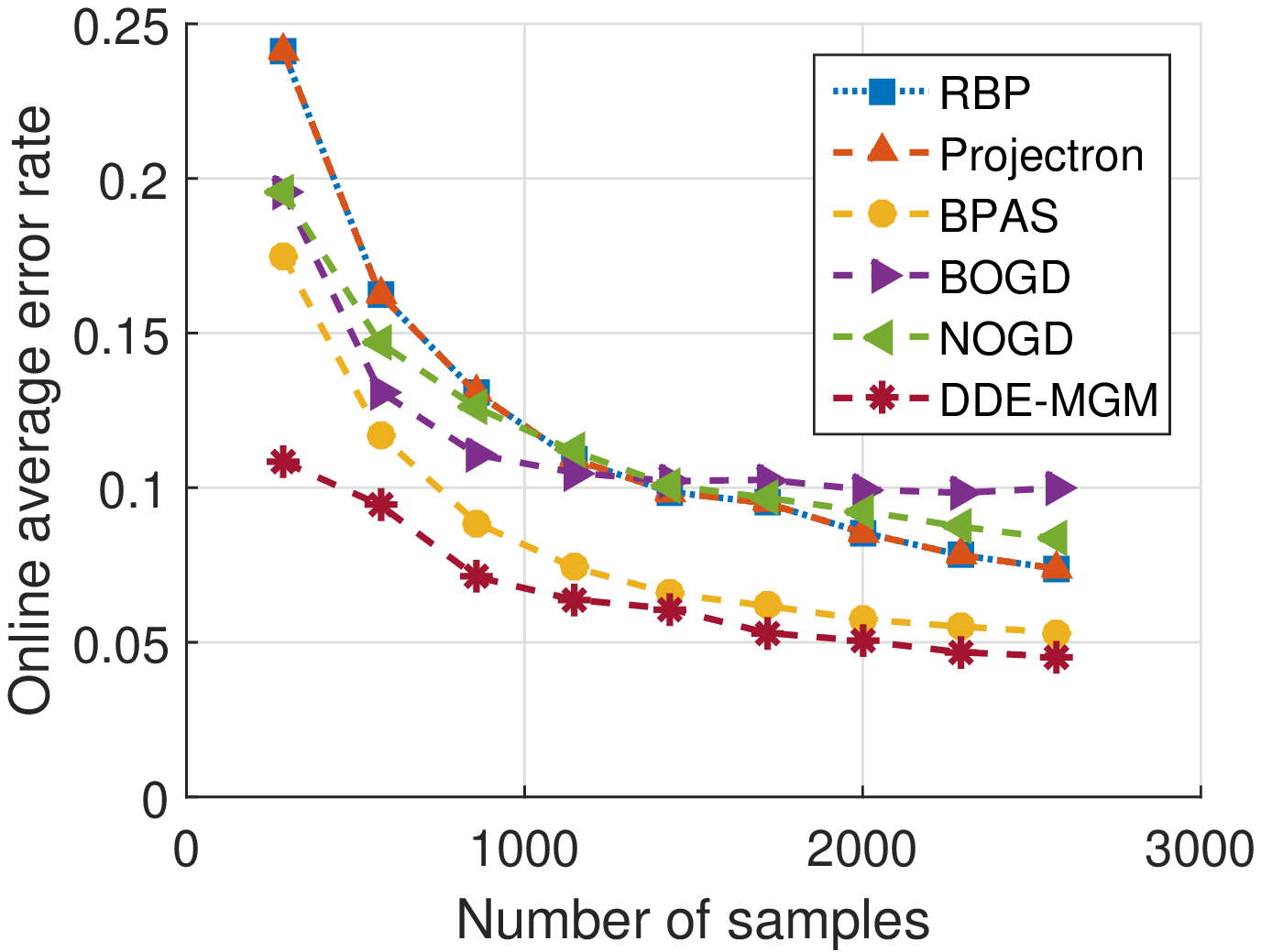}\label{subfig:UCI_online_error}}
	\subfigure{\includegraphics[width=.48\columnwidth]{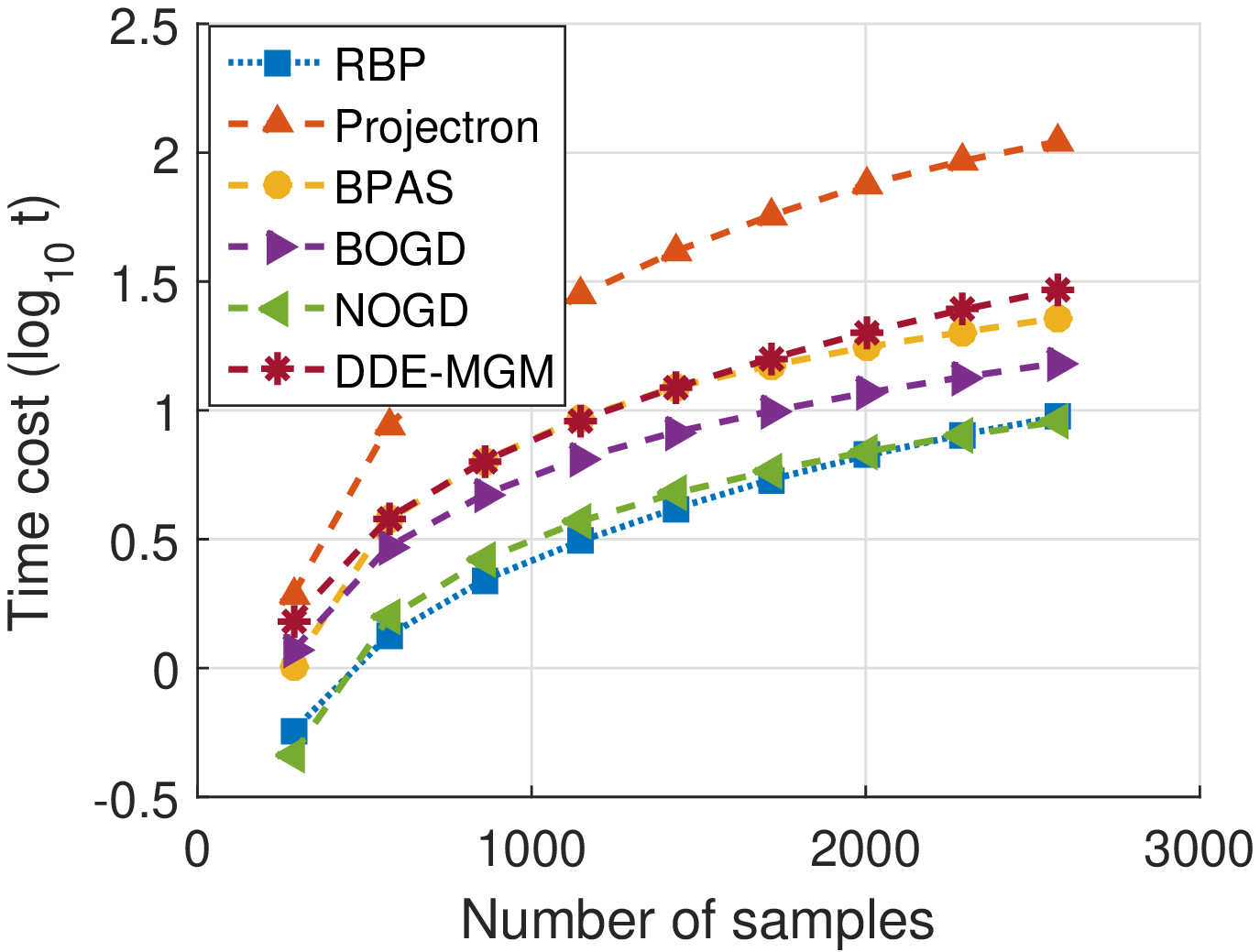}\label{subfig:UCI_online_time}}
	\caption{Comparison of online algorithms on the UCI character trajectories dataset.}
	\label{fig:UCI_online}
\end{figure}

In the online experiment, although DDE-MGM is not the most efficient, it achieves the highest accuracy, even in the earlier stage (fewer samples) as illustrated in Fig.~\ref{fig:UCI_online}. 

\textbf{The MSR Action3D dataset:}
The most notable advantage of DDE-MGM over the other online algorithms is the robustness to random data length and misalignment, which is better demonstrated in the experiment on the MSR Action3D dataset as shown in Table~\ref{tab:MSR_cmp}. For this dataset, DDE-MGM significantly outperforms the other algorithms in both off- and on-line testing because the raw data is not well aligned and varies in length. 
\begin{table}[h]
	\centering
	\caption{Comparison on the MSR Action3D dataset}
	\begin{tabular}{c|c|c|c}
		\hline\hline
		Algorithm & Accu. (\%) & Time (sec) & O / R\\
		\hline
		1NN-DTW & 74.73 & 7.6$\times 10^4$ & - / +\\
		SAX & 61.90 & 54.68  &   - / -\\
		HMM & 60.07 & 2.1$\times 10^3$ &   - / -\\
		DDE-MGM & \textbf{93.04} & \textbf{28.40} & + / + \\
		\hline
		RBP & 23.41 & 20.23 & + / - \\
		Projectron & 31.65 & 205.25 & + / -\\
		BPAS & 30.36 & 12.25 & + / -\\
		BOGD & 26.19 & 22.23 & + / -\\
		NOGD & 29.96 & \textbf{10.47} & + / -\\
		DDE-MGM & \textbf{79.37} & 80.38 & + / + \\
		\hline\hline
	\end{tabular}
	\label{tab:MSR_cmp}
\end{table}

\begin{figure}[h]
	\centering
	\subfigure{\includegraphics[width=.48\columnwidth]{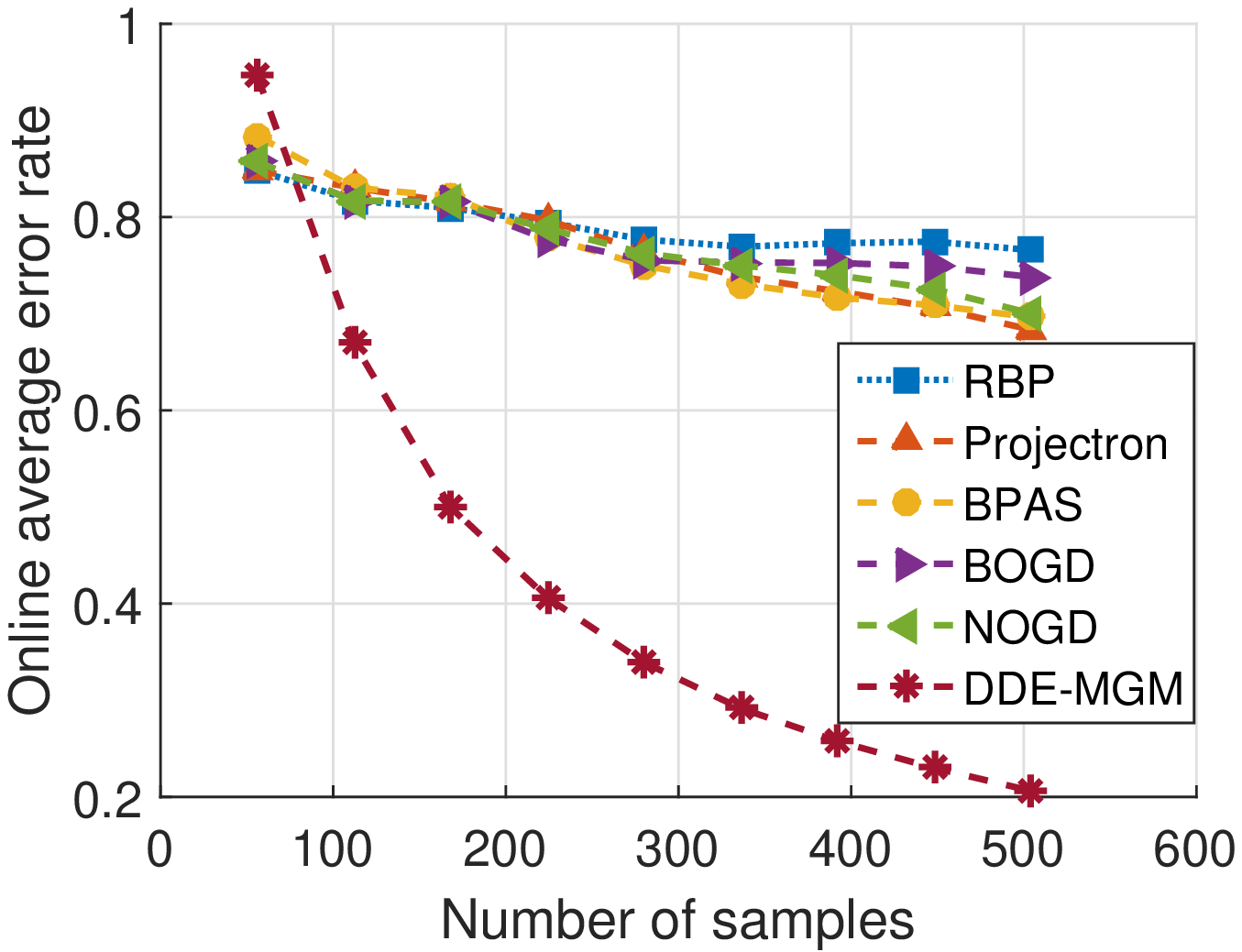}\label{subfig:MSR_online_error}}
	\subfigure{\includegraphics[width=.48\columnwidth]{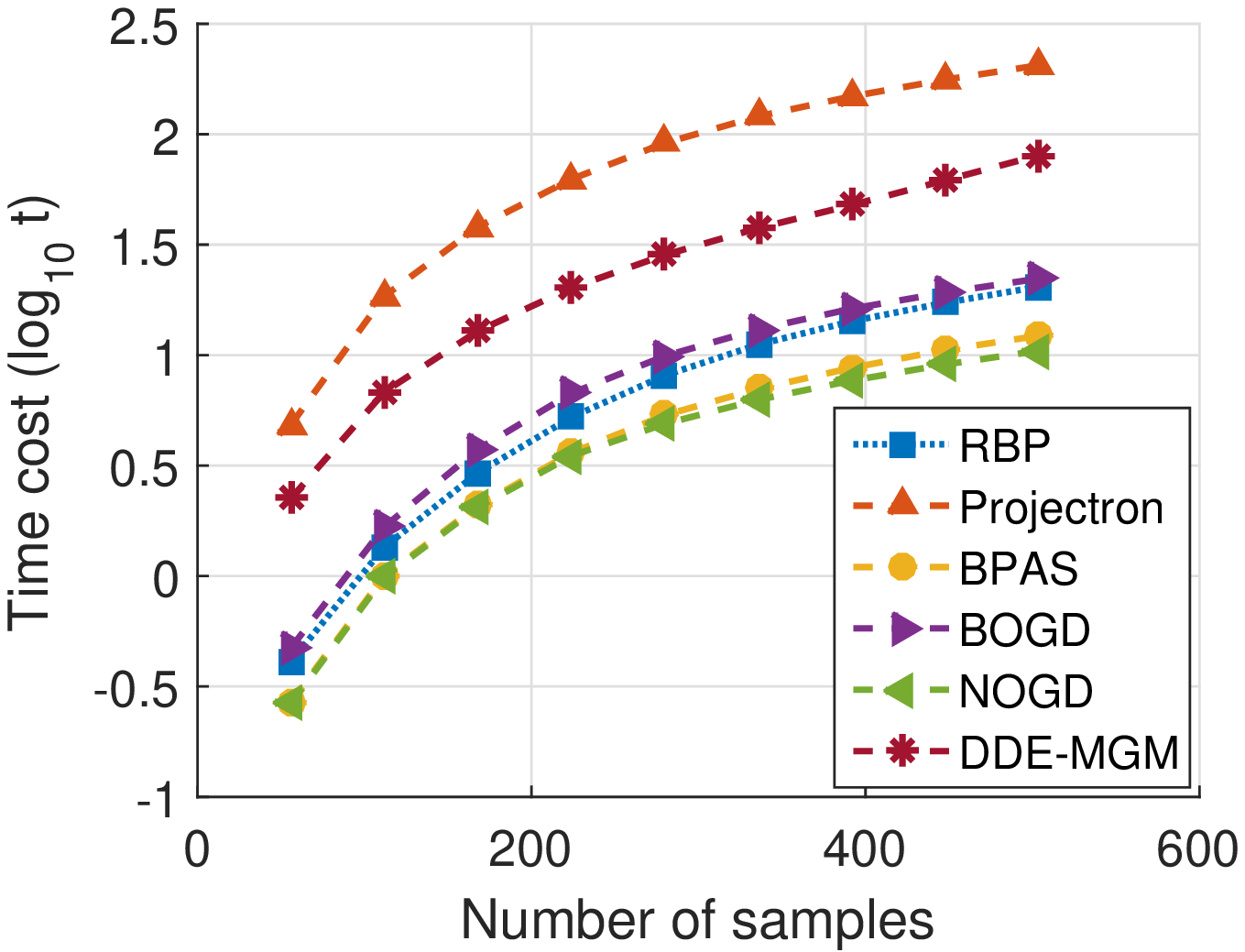}\label{subfig:MSR_online_time}}
	\caption{Comparison of different online algorithms on the MSR Action3D dataset}
	\label{fig:MSR_online}
\end{figure}

The highest accuracy of the other online algorithms is around 30\% (BPAS) because they are sensitive to the alignment of the time series. Fig.~\ref{fig:MSR_online} compares the online performance where it is obvious that DDE-MGM still preserves relatively good performance.

Because MSR Action3D is one of the most popular datasets used by the action recognition community, we follow the same experimental setup in most related works for a fair comparison with the state-of-the-art performance. The accuracy we obtain is about 93\%, and the literature~\cite{vemulapalli2014human,anirudh2015elastic} published in recent years achieved around 90\%. Therefore, DDE-MGM is still competitive to those algorithms specifically designed for this dataset. 

\textbf{The PAMAP dataset:}
Actually, the previous two datasets are not infinite streaming time series because the time duration is only a few seconds for each sample. Therefore, all the cited algorithms are modeling on multiple examples (segments) rather than on a data stream. For a streaming time series, there is not specific start or end time, so that the online algorithms cited in this paper cannot work without employing extra segmentation methods. After incorporating certain segmentation algorithm, however, the algorithms may loose the online property or yield lower classification accuracy. 
To demonstrate the effectiveness of DDE-MGM on modeling streaming time series, the PAMAP dataset is adopted because its samples last tens of minutes (comprising tens of thousands of points) that could be considered a streaming time series. Assuming the data points arrive one-by-one, DDE-MGM incrementally models the stream without segmentation or any other preprocessing. Table~\ref{tab:PAMAP_cmp} reports the results of DDE-MGM, as well as the offline methods. Leave-50\%-out cross validation is applied, and the samples are truncated into the same length for SAX and HMM. The online methods are not involved in this comparison because they require extra segmentation algorithms.
\begin{table}[h]
	\centering
	\caption{Comparison on the PAMAP dataset}
	\begin{tabular}{c|c|c|c}
		\hline\hline
		Algorithm & Accu. (\%) & Time (sec) & O / R\\
		\hline
		1NN-DTW & 69.57 & 1.44$\times^4$ & - / +\\
		SAX & 56.52 & 161.81  &   - / -\\
		HMM & 60.87 & 5.2$\times 10^3$  &   - / -\\
		Dictionary & 84.80 & 7.9$\times 10^3$ & - / + \\
		DDE-MGM & \textbf{86.96} & \textbf{135.86} & + / + \\
		\hline\hline
	\end{tabular}
	\label{tab:PAMAP_cmp}
\end{table}

The ``Dictionary'' denotes the algorithm in~\cite{hu2013time}, which relaxed the fixed-length assumption by learning a dictionary in an offline manner. It achieves the state-of-the-art accuracy of 84.8\% that is a bit lower than DDE-MGM, but its run time is drastically longer than the proposed. 

\subsection{How Fast is DDE-MGM Model?}

We have claimed that DDE-MGM can achieve online modeling. However, no online method can handle a time series with infinite streaming rate. So, what is the limit of DDE-MGM? To find the limitation, we use the PAMAP dataset again because its long time duration is suitable to examine the maximum modeling speed. 

In the modeling stage of DDE-MGM, there are totally three parameters --- delay step $s$, embedding dimension $d$ and cell size of the discretized embedding space. The parameters $d$ and $s$ determine how many points and what interval they are extracted from the data stream to reconstruct a state in the embedding space. Variation of these parameters will not affect the computational complexity of DDE, whose run time is approximately constant. Therefore, we may ignore the effect of $s$ and $d$ on modeling speed. The only parameter remained is the cell size, which significantly affects the modeling speed in our experiment. A smaller cell size, for example,  will result in more cells in the discretized embedding space, so that the grid size (the number of bins on each dimension) will be larger, and more states and transitions need to be recorded. Note that the cell size is the size of a cell in the grid, and the grid size refers to the number of cells on each dimension of the grid.

As discussed in section~\ref{subsec:online_modeling}, we use a ``list'' to represent the sparse transitions, therefore a larger grid size generates a longer list. Most run time of DDE-MGM during modeling is consumed on searching the list for accumulating new transitions to existing ones, so larger grid size results in longer run time as shown in Table~\ref{tab:accuracy2}. In addition, larger grid size does not necessarily increase the accuracy because when the grid size goes larger (i.e., the cell size goes smaller), the overfitting problem starts to surface and the model ends up fitting noisy data. 
The grid size of 50 is an appropriate setting based on extensive empirical studies.
\begin{table}[h]
	\centering
	\caption{Efficiency of DDE-MGM (experiments on the PAMAP dataset)}
	\begin{tabular}{r|ccccc}
		\hline\hline
		Grid size & 20 & 30 & 40 & \textbf{50} & 60 \\\hline
		Accu. (\%) & 56.7 & 70.9 & 79.1 & \textbf{86.9} & 85.0 \\\hline
		Time (sec) & 15.5 & 29.1 & 65.9 & \textbf{135.8} & 166.9  \\\hline
		Memory (KB)  & 2 & 3 & 5 & \textbf{7} & 9 \\\hline
		Rate (kHz) & 13.1 & 12.7 & 12.1 & \textbf{11.6} & 11.3\\
		\hline\hline
	\end{tabular}
	\label{tab:accuracy2}
\end{table}

The efficiency in the aspect of memory cost is uniquely determined by the grid size. Also shown in Table~\ref{tab:accuracy2}, the memory cost increases monotonously with the grid size. The raw time series from one class is over 10MB, the memory footprint of the learned model from one class is less than 7KB under the grid size of 50. Because the grid size is fixed, the number of memory units is fixed as well; then the memory cost approaches a constant regardless of the stream length.

To explicitly show how fast DDE-MGM can model a streaming time series, we investigate the maximally-allowed streaming rate --- the maximum points from the time series that can be updated to the MGM model in one second, as shown in the last row of Table~\ref{tab:accuracy2}. The results are obtained by modeling several randomly truncated time series of 10,000 points from the PAMAP dataset. When the grid size is 50, for example, the averaged run time on each truncated time series is about 0.86 sec, thus the maximally-allowed streaming rate is $10,000/0.86\approx 11,600$ Hz, which is sufficient for most real-world applications. Note that the maximally-allowed streaming rate only varies with the grid size. 

\subsection{Effect of Parameter Setting}
\begin{figure}[h]
	\centering
	\subfigure[UCI dataset]{\includegraphics[width=.48\columnwidth]{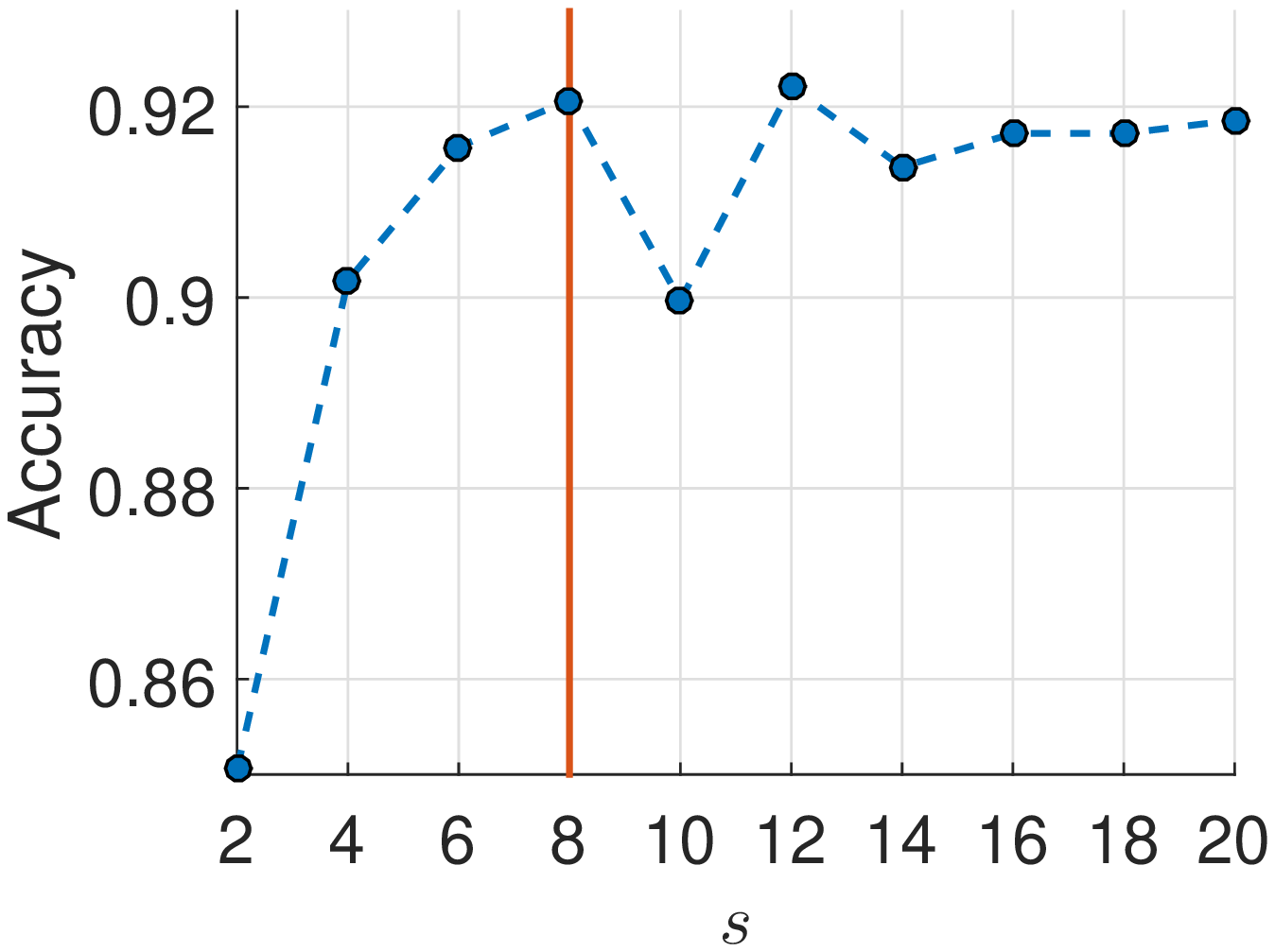}\label{subfig:s_vs_accu_a}}
	\subfigure[MSR Action3D]{\includegraphics[width=.48\columnwidth]{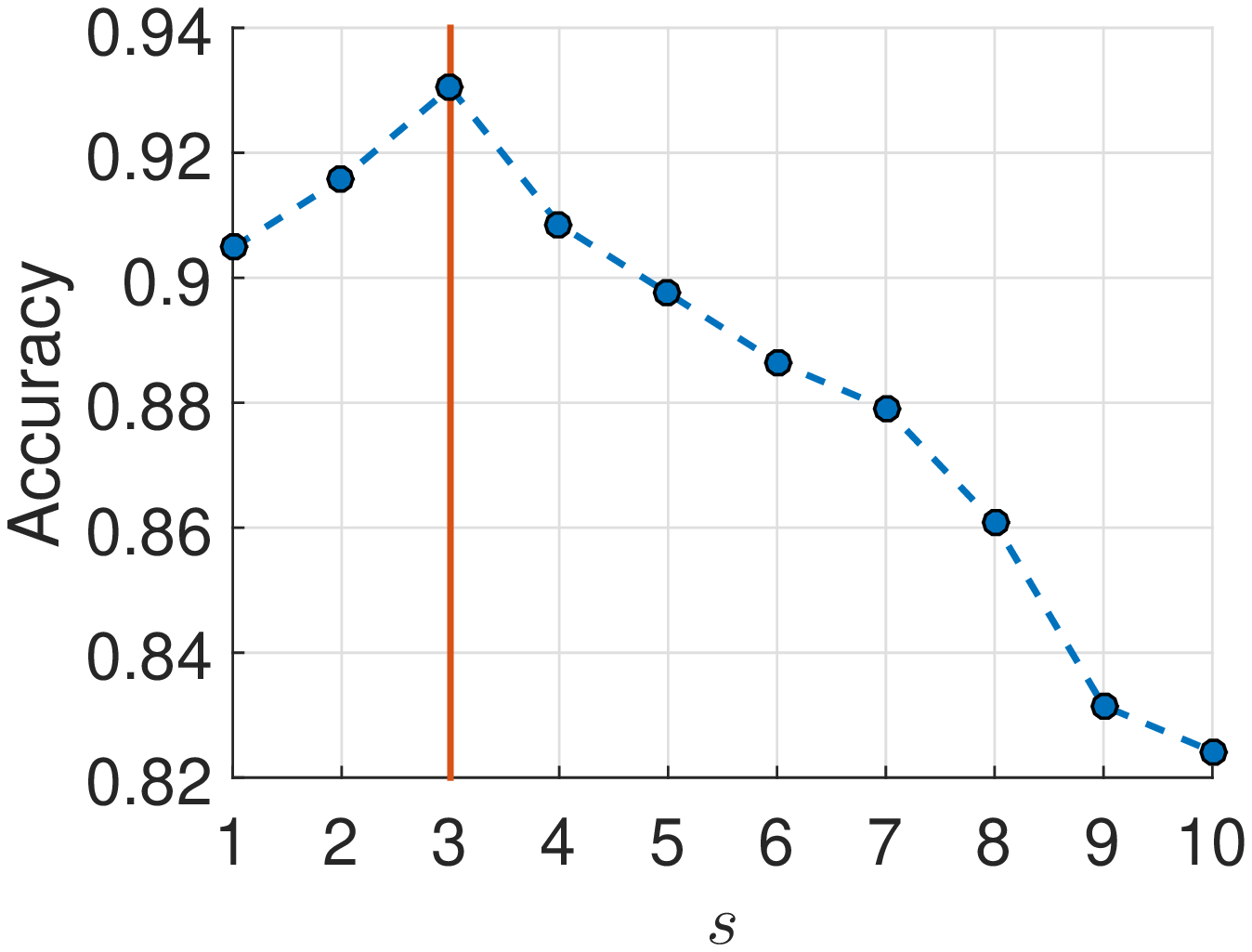}\label{subfig:s_vs_accu_b}}
	\caption{Effect of delay step $s$ on classification accuracy. The solid vertical lines denote the parameters selected in our experiment.}
	\label{fig:s_vs_accu}
\end{figure}

Although appropriate parameter settings for the delay step $s$ and embedding dimension $d$ can be obtained based on the methods in \cite{perea2013sliding} and \cite{kennel1992determining}, respectively, it is still interesting to see the effect of the two parameters on classification accuracy. 
This section compares the accuracy using different $s$ or $d$ on the UCI character trajectories and MSR Action3D datasets. Fig.~\ref{fig:s_vs_accu} demonstrates the effect of $s$ on classification accuracy.

\begin{figure}[h]
	\centering
	\subfigure[UCI dataset]{\includegraphics[width=.45\columnwidth]{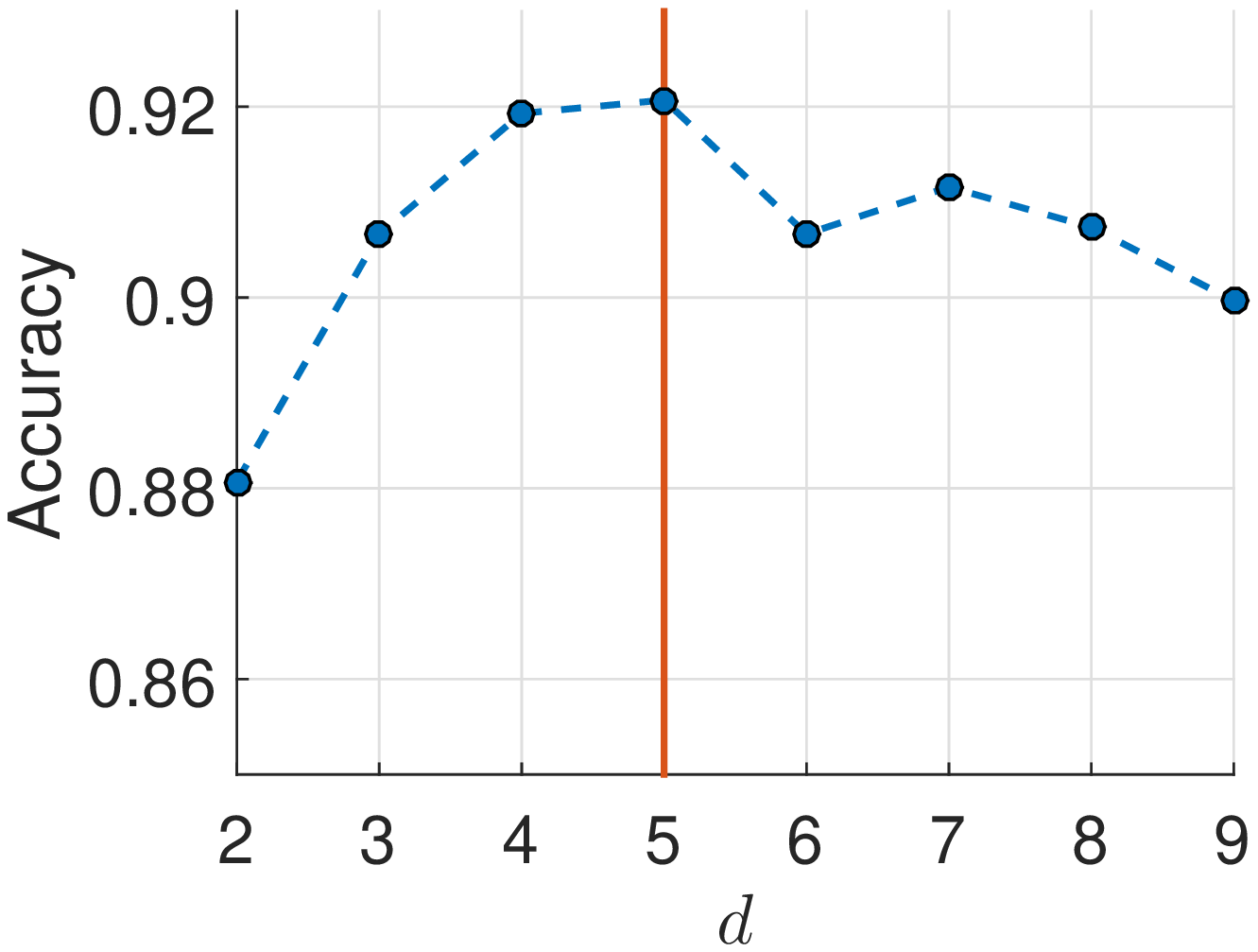}\label{subfig:d_vs_accu_a}}
	\subfigure[MSR Action3D]{\includegraphics[width=.45\columnwidth]{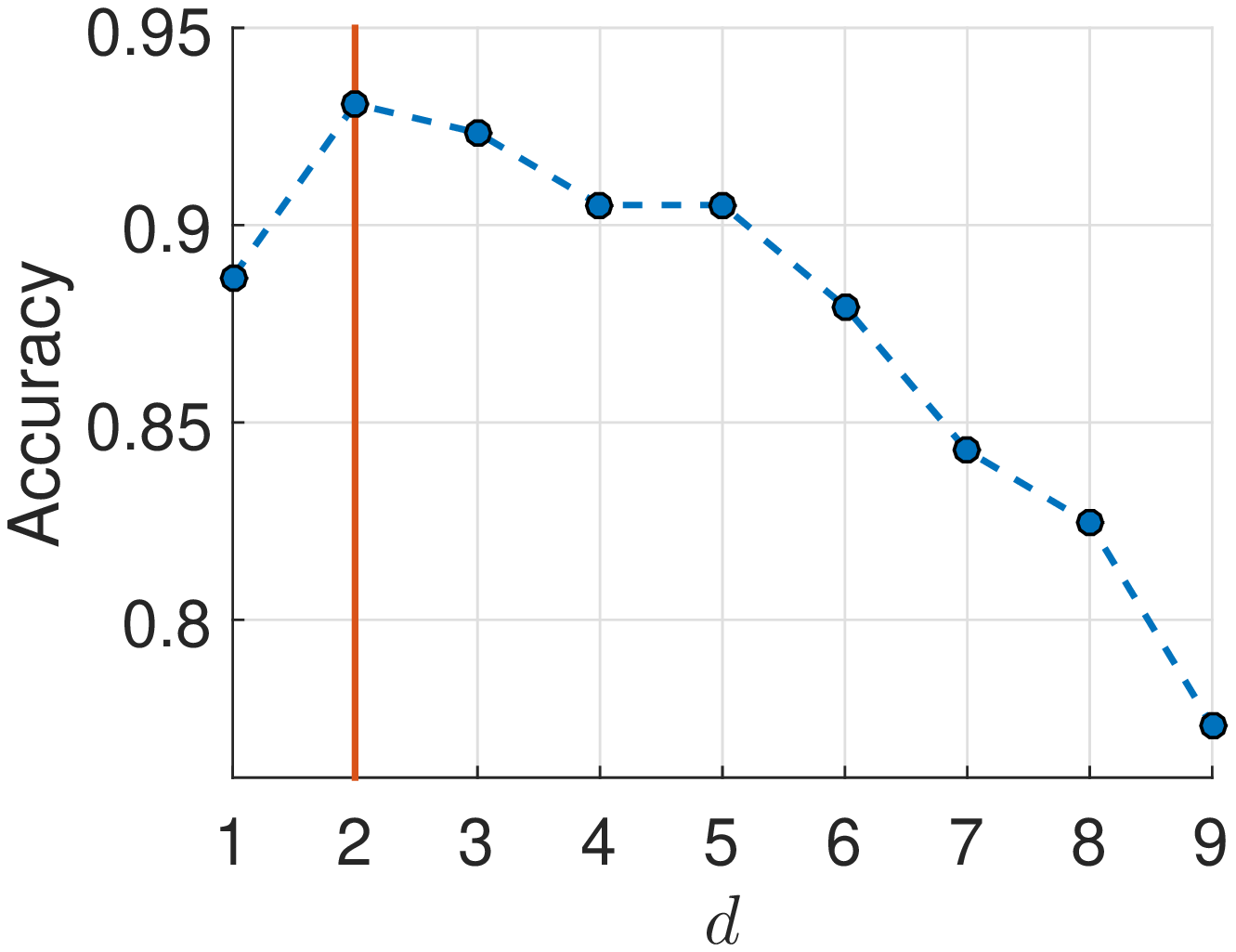}\label{subfig:d_vs_accu_b}}
	\caption{Effect of embedding dimension $d$ on classification accuracy. The solid vertical lines denote the parameters selected in our experiment.}
	\label{fig:d_vs_accu}
\end{figure}

From Fig.~\ref{subfig:s_vs_accu_a}, the selected $s$ in our experiment based on the method in \cite{perea2013sliding} is not necessarily the optimal $s$ because the accuracy is a little bit higher when $s=12$ (92.21\%). However, the accuracy of selected $s$ is very close to that of the optimal $s$. By the same token, the selected $d$ is not necessarily the best in general as shown in Fig.~\ref{fig:d_vs_accu}.

Above all, both $s$ and $d$ significantly affect the classification accuracy. The extent of effect mainly depends on the dataset. If the dominant frequency of the samples is small, the changes of $s$ and $d$ will cause less variation on the accuracy, and vice versa. For example, the samples in the UCI dataset have lower dominant frequency than that in the MSR Action3D dataset, so the changing of parameters affects more on the latter. If the parameters vary around the selected values (deviating 1 or 2 from the selected value), the accuracy changes about two percent for the UCI dataset. In contrast, the accuracy changes approximately four percent on the MSR Action3D dataset.    

\section{Conclusions}
\label{sec:Conclusions and Future Work}
In this paper, we proposed a novel method, DDE-MGM, to model and classify time series in an online manner, where common but unrealistic assumptions like the same data length and well alignment are completely removed, facilitating the deployment of the method to real-world problem solving. The main objective of DDE-MGM is computational efficiency from the aspects of both computing time and memory consumption, while preserving superior classification accuracy as compared to the state-of-the-art methods. The experiments conducted on three real datasets had validated (1) the effectiveness of using the trajectory in the embedding space to distinguish the intrinsic patterns of different classes in the time-domain training steam, (2) the flexibility and feasibility of the novel online processing scheme for streaming data without making any assumptions, (3) the great potential for modeling and classification in real time, and (4) the small and constant memory footprint. 


\bibliographystyle{abbrv}
\bibliography{CIKM2016References}

\end{document}